\documentclass[journal]{IEEEtran}

\usepackage{amsmath}
\usepackage{amsfonts}
\usepackage{algorithm}
\usepackage[noend]{algpseudocode}
\usepackage{algorithm}
\usepackage{booktabs}
\usepackage{array}
\usepackage[numbers]{natbib}
\usepackage{multirow}
\usepackage[labelformat=simple]{subcaption}
\usepackage{graphicx}
\usepackage{listings}
\usepackage[utf8]{inputenc}
\usepackage[T1]{fontenc}
\usepackage{fancyvrb}
\usepackage[final]{changes}
\newcommand{\stkout}[1]{\ifmmode\text{\sout{\ensuremath{#1}}}\else\sout{#1}\fi}
\setdeletedmarkup{\stkout{#1}}

\begin{document}
\bstctlcite{IEEEexample:BSTcontrol}

\title{Automated Video Game Testing Using\\Synthetic and Human-Like Agents}

\author{
    \IEEEauthorblockN{Sinan~Ariyurek\IEEEauthorrefmark{0},~Aysu~Betin-Can\IEEEauthorrefmark{0},~Elif~Surer\IEEEauthorrefmark{0}}
    \IEEEauthorblockA{\IEEEauthorrefmark{0}
    \\Graduate School of Informatics
    \\Middle East Technical University
    \\06800, Ankara, Turkey
    \\\{sinan.ariyurek, betincan, elifs\}@metu.edu.tr}
}

% \markboth{IEEE TRANSACTIONS ON GAMES, VOL. X, NO. X, 20XX}{}%
\markboth{}{}%

\maketitle

\begin{abstract}
In this paper, we present a new methodology that employs tester agents to automate video game testing.\deleted{These agents are analogous to game-playing agents but focus on finding defects.} We introduce two types of agents ---synthetic and human-like--- and two distinct approaches to create them. \added{Our agents are derived from Reinforcement Learning (RL) and Monte Carlo Tree Search (MCTS) agents, but focus on finding defects.} \deleted{Former harnesses game scenarios to generate test goals and further modifies them to examine the effect of unintended game transitions.} \added{The synthetic agent uses test goals generated from game scenarios, and these goals are further modified to examine the effects of unintended game transitions.} \deleted{The latter uses our proposed multiple greedy-policy inverse reinforcement learning (MGP-IRL) to extract test goals from tester trajectories.} \added{The human-like agent uses test goals extracted by our proposed multiple greedy-policy inverse reinforcement learning (MGP-IRL) algorithm from tester trajectories.} MGP-IRL captures multiple policies executed by human testers. \replaced{These}{Note that, a} tester\replaced{s'}{'s} aim\added{s} \replaced{are}{is} finding defects \replaced{while}{and} interacting with the game to break it, which is considerably different from game playing. We present interaction states to model such interactions.\deleted{Our agents use Reinforcement Learning (RL) and Monte Carlo Tree Search (MCTS) to produce test sequences,} \added{We use our agents to produce test sequences,} run the game with these sequences, and check the game for each run with an automated test oracle. We analyze the proposed method in two parts: we compare the success of human-like and synthetic agents in bug finding, and we evaluate the similarity between human-like agents and human testers. We collected 427 trajectories from human testers using the General Video Game Artificial Intelligence (GVG-AI) framework and created three \deleted{test bed} games with 12 levels that contain \deleted{40}\added{45} bugs. Our experiments reveal that human-like and synthetic agent\added{s} compete with human testers' bug finding performances. Moreover, we show that MGP-IRL increases the human-likeness of\deleted{ the} agent\added{s} while improving the bug finding performance.
\end{abstract}

\begin{IEEEkeywords}
Reinforcement Learning, Monte Carlo Tree Search, Automated Game Testing, Inverse Reinforcement Learning, Graph Coverage.
\end{IEEEkeywords}

\IEEEpeerreviewmaketitle

\section{Introduction}\label{sec:introduction}

Video games industry is a multi-billion industry that is continually growing \cite{Lin:2019}. Though the success of a video game depends upon numerous aspects, the presence of bothersome bugs decrease\added{s} the overall experience of a player. Moreover, the bugs that are found after release not only increase the overall budget \cite{Boehm:1988}, but also act as a negative feedback on the development and testing team. Hence, the game is tested painstakingly by game developers and players, and these tests require immense tester effort. The major difficulty of game testing arises from the constant changes \cite{Santos:2018}\deleted{, and an alteration to the game design demands tests to be repeated; thus a flexible approach is required}. Therefore, researchers proposed \added{various methods to decrease the test effort. These methods are} regression tests based on record/replay segments \cite{Ostrowski:2013}, scenario testing \cite{Cho:2010}, UML\added{-}based sequence generation \cite{Iftikhar:2015}, agents harnessing Petri nets \cite{Becares:2016} and RL agents \cite{Pfau:2017}\cite{Loubos:2018} to automate this process. \replaced{These studies, however,}{Although these methods are used to automate testing in a single domain, they} lack either human expertise, an automated oracle, an intelligent tester agent, or an overall game testing experiment.

In game playing, researchers employed artificial intelligence (AI) to make agents behave like human beings from collected human data\added{, and called these agents as human-like agents in the literature}. Human-like agents are better suited to analyze the difficulty of the game \cite{Gudmundsson:2018}, can become genuine opponents \cite{Tastan:2011}\cite{Glavin:2015}, and can generate satisfying playthroughs. In game testing, these collected human data are used to perform regression testing rather than creating an intelligent agent. During alpha and beta testing phases \cite{Redavid:2011}, countless test data can be collected from players. Human game testers participating in these phases use their expertise\deleted{ and instincts} to examine the game. We propose a method to capture this expertise \added{in the form of test goals} and \deleted{generate}\added{use these test goals in} agents \added{so} that \added{they can} test like the original human player. \added{Test goals are objectives that agents want to validate in a game. Test goals range from whether the game can be finished to whether the agent can walk through walls. Depending on the test goal, agents generate a different test sequence.}\deleted{Therefore, these agents have an advantage over regression testing as they can also be used to test other levels.} In this paper, \deleted{it}\added{the goals that are extracted from collected human data are called human-like test goals.}\deleted{ is referred to as the human-like agent.}

% \added{Since these goals are directly learned through human testers, we call them human-like test goals. These test goals can be used by agents to generate test sequences for other levels that have a different layout.}

\begin{figure}[]
  \centering
  \includegraphics[width=0.8\columnwidth]{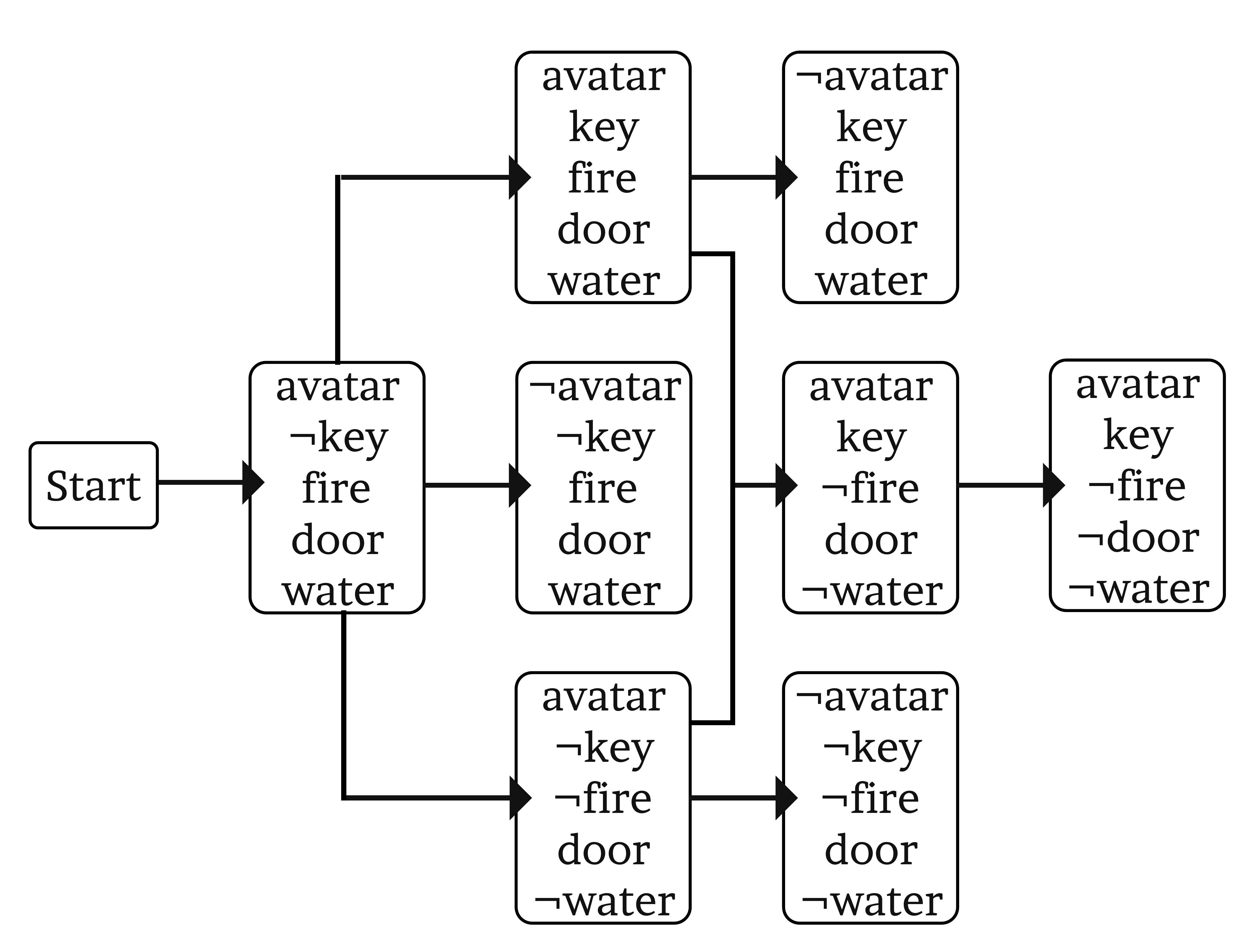}
  \caption{A scenario graph model using atomic properties}
  \label{fig:scn_graph}
\end{figure}

\begin{figure}[!ht]
  \centering
  \includegraphics[width=0.96\textwidth]{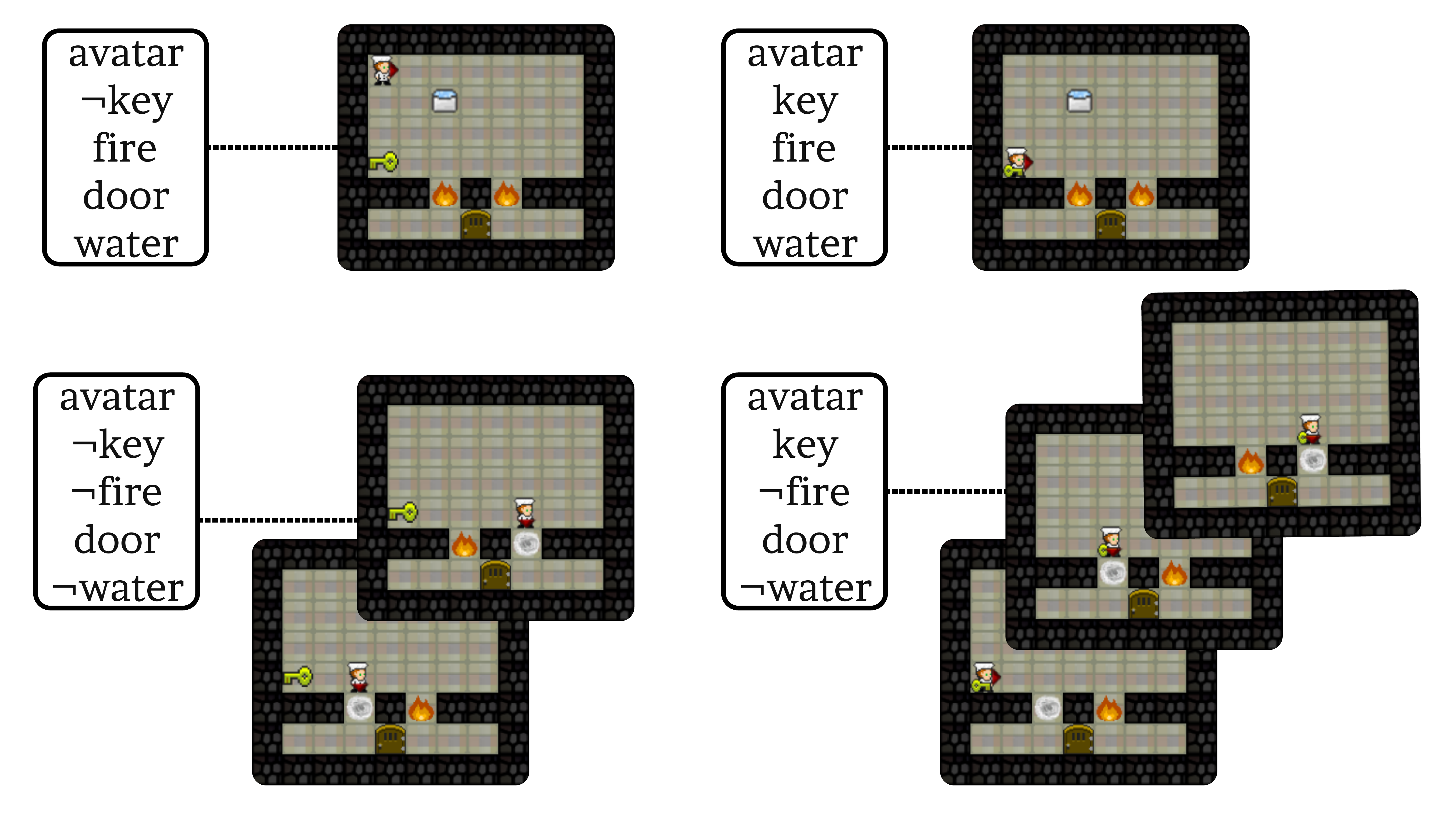}
  \caption{Realized Nodes of Game Scenario Graph from \figurename{} \ref{fig:scn_graph}.}
  \label{fig:scn_realized_graph}
\end{figure}

On the other hand, a game can be viewed as the implementation of the game designer's story. This story ---whether linear or non-linear--- can be represented using a graph \cite{Adams:2014}. In this paper, this graph is referred to as a game scenario graph. \added{Game scenario graph (see \figurename{ \ref{fig:scn_graph}}) is designed by the game designer and contains high-level behavior.} A node on this graph is \replaced{realized}{correlated} with \added{the} states of the game \added{(see \appendixname{ \ref{sec:appendix_1}} \figurename{ \ref{fig:scn_realized_graph}})}. Edges are the actions that progress the story. Additionally, as directed graphs form the foundation of several coverage criteria in software testing \cite{Ammann:2008}, it is possible to generate test sequences as paths using this graph and a coverage criterion. We enhance \deleted{this method with extending }the test sequence by adding actions at each \replaced{node}{state} that should not progress the game. The \replaced{criterion-based paths}{former technique} \replaced{verify}{verifies} the implementation of the game scenario while the \replaced{enhancements}{latter} check\deleted{s} other aspects of the game such as testing\deleted{ the} collisions\deleted{,} and unintended actions. We propose a method that translates these ideas \added{in}to \added{test goals. Since no human data are used, we call these goals synthetic test goals.}\deleted{ an agent since no human data are used, this agent is called the synthetic agent.}
 
Game researchers used RL \added{agents }to play various games such as Ms. Pac-Man \cite{Tziortziotis:2014}, Bomberman \cite{Kormelink:2018}, Unreal Tournament \cite{Tastan:2011}\cite{Glavin:2015}\replaced{. Furthermore,}{, and} recent developments in AI showed that agents can surpass humans in \replaced{arcade games \cite{Mnih:2015}, Go \cite{Silver:2016} and StarCraft II \cite{AlphaStar:2019}}{StarCraft II \cite{AlphaStar:2019}, Go \cite{Silver:2016} and arcade games \cite{Mnih:2015}}. The success in Go \cite{Silver:2016} \replaced{is achieved with RL, Supervised Learning, and MCTS}{can be apportioned to MCTS}. Moreover, MCTS \cite{Browne:2012} \added{agents are}\deleted{is} found to be successful on GVG-AI \cite{GVGAI:2019} and \added{General Game Playing (}GGP\added{)} \cite{GGP:2014}, which are the most well-known frameworks that explore agents that can play various games. \added{In our system, MCTS and RL agents use the synthetic or human-like test goals}\deleted{Therefore, our tester agents use MCTS and RL, but separately} \added{to generate test sequences}. \deleted{These algorithms generate the desired sequence given a synthetic or human-like agent.} \added{In this paper, a synthetic agent is an RL or an MCTS agent which uses synthetic test goals, and a human-like agent is an RL or an MCTS agent which uses the extracted human test goals.}

% \begin{figure}[]
%   \centering
%   \includegraphics[width=1.0\columnwidth]{ariyu1_revision.pdf}
%   \caption{An overview of the system}
%   \label{fig:system_overview}
% \end{figure}

% This paper presents a framework that generates \replaced{test goals for}{intelligent} agents to automate video game testing. \deleted{Fig. \ref{fig:system_overview} illustrates our proposed system. }Two models of \deleted{agents}\added{test goals} are proposed, human-like and synthetic\deleted{ (as seen on the top part of \figurename{ \ref{fig:system_overview}})}. Human-like \deleted{agents}\added{test goals} are generated from the collected human tester trajectories. Synthetic \deleted{agents}\added{test goals} are produced from the game scenario graph. 

\deleted{This paper presents a framework that generates intelligent agents to automate video game testing. Fig. \ref{fig:system_overview} illustrates our proposed system. Two models of agents are proposed, human-like and synthetic (as seen on the top part of \figurename{ \ref{fig:system_overview}}). Human-like agents are generated from the collected human tester trajectories. Synthetic agents are produced from the game scenario graph. Using these agents, new test sequences are generated for a game under test using RL or MCTS (see bottom-left part of Fig. \ref{fig:system_overview}). These generated sequences are replayed and checked by an automated test oracle. }\added{Current off-the-shelf automated testing tools help testers to implement and automate test scenario executions. However, these tools do not design test scenarios or test sequences. Our study fills this gap with intelligent agents. Our synthetic and human-like agents generate test sequences for a game under test without human intervention. Then, our framework replays these generated test sequences and check the game behavior with an automated oracle. }The test oracle checks if the game behaves as expected according to the game design \replaced{constraints}{rules}\added{ and game scenario graph which are given by the game developer}. Our aim is to find discrepancies between the actual implementation and the design of the game, thus visual glitches are not checked by our oracle. We present our approach using the GVG-AI framework. GVG-AI is chosen since it contains numerous games, \added{and }has a data collection mechanism\deleted{, and bugs can be inserted using VGDL \cite{VGDL:2014}}. In this paper, our approach is \replaced{demonstrated via}{shown using} grid games, but it can be generalized to different game models.

% We consider VGDL as the implementation and the game design as the ideal bug-free concept of a game. The game scenario graph and the rules match the game design.

Our first contribution is human-like tester agents, to the best of our knowledge, our work is the first to propose these agents in game testing. The proposed human-like agents can be used to test other levels, thus possess an advantage over \replaced{record/replay}{regression} testing. Second, the synthetic agent is an improvement over simple scenario testing, rewarding the examination of all allowed transitions and some unallowed game transitions for robustness. Lastly, we present an interaction state that enable\added{s} us to capture tester \replaced{strategies}{instincts} and play them.

This paper is structured as follows: Section \ref{sec:preliminaries} gives preliminary information about Graph Testing, RL, MCTS and GVG-AI. Section \ref{sec:related_research} describes the examples and methodologies of related research. Our proposed methodology is presented in Section \ref{sec:methodology} while the details of our experiments are illustrated in Section \ref{sec:experiments}. Section \ref{sec:discussion} discusses the outcome\added{s} of the strategies used, their contributions and limitations. Section \ref{sec:conclusion} concludes this paper.

\section{Preliminaries}\label{sec:preliminaries}

The following subsections introduce the preliminary material\deleted{, outlined as follows}: Graph Testing, RL, MCTS, \added{and} GVG-AI.

\subsection{Graph Testing}

In software testing, there are several systematic ways to create tests and evaluate the adequacy of the test set. Testers model the system under test (SUT) and check for defects with the help of this model. One common way to model software is using graphs.\deleted{ Some examples are program dependency graphs, control flow graphs, data flow graphs, directed graphs representing the transitions between multiple screens, statecharts and so on.} Once we model the SUT with a graph, we can apply systematic testing techniques to generate test scenarios\deleted{ and evaluate the adequacy of the test set}.

A directed graph is a structure that consists of nodes $N$ and edges $E$ where $E \subseteq N \times N$. A path \replaced{is}{can be written as} a sequence $[n_1,n_2,...,n_M]$ of nodes where each pair is in the set of edges $E$. For graphs, various sequences can be obtained depending on the coverage criterion. The coverage definitions are stated from basic to complex. \textbf{Edge Coverage (EC):} Contains each reachable path of length up to one. \textbf{Edge-Pair Coverage (EPC):} Contains each reachable path of length up to two. \textbf{Simple Path (SP):} Requires a node not to appear more than once, unless it is the initial or final node of the path. \textbf{Prime Path (PP):} Strengthens \added{the} simple path definition by requiring the path not to be a sub-path of another simple path. \textbf{Prime Path Coverage (PPC):} Contains each prime path in the graph. \textbf{All Path Coverage (APC):} Contains every path in the graph. The order of coverage is: $EC \subseteq EPC \subseteq PPC \subseteq APC$ \cite{Ammann:2008}.

\subsection{Reinforcement Learning} \label{sec:pre:rl}

% Section II.B. "The optimal policy can be found using Q-Learning." - is Q-Learning the only way to find the optimal policy? This whole section seems a bit messy, throwing in several concepts without any proper introduction or reasoning as to why they are even there in the first place, and lacking proper explanations.

% Important details of the RL algorithm configurations are missing in Section IV.C Generating Test Sequences. Specifically, the paper lacks a clear definition of the state representation, action space and policy used in the MCTS and SARSA-lambda.

% Reinforcement learning section has some symbols that are not defined and being used.

\deleted{
Reinforcement learning (RL) is experiencing an environment, examining its current state and learning how to interact with this environment by use of possible actions and reward mechanisms. This decision model can be written using the Markov Decision Process (MDP) \cite{Sutton:2018}.
}

\added{
In reinforcement learning (RL), an agent experiences an environment, examines its current state and interacts with it. The agent receives a reward and a new state from the environment as a response to these interactions. Markov Decision Process (MDP) defines this experience between an agent and the environment \cite{Sutton:2018}, and the RL problem can be written using an MDP.} Markov Decision Process is a tuple $(S, A, T, R, \gamma)$ where $S$ is the set of states, $A$ is the set of actions, \added{and }$T: S \times A \times S \rightarrow [0,1]$ is the transition probability\deleted{ that defines $T(s, a, s')$ where $s, s' \in S$, $a \in A$}. \added{The reward function is $R: S \times A \rightarrow \mathbb{Q}$, and $\gamma$ is the discount rate for future rewards.}

\deleted{
The objective is to find a policy $\pi : S \rightarrow A$, basically telling which action to take, given a state. The optimal policy can be found using \textit{Q-Learning}. The parameter $\alpha$ controls the update amount.
}

\begin{equation} \label{eq:sarsa}
  \begin{aligned}
  % & \deleted{Q(s,a) \gets Q(s,a) + \alpha [R(s,a) + \gamma \max_{a'} Q(s', a') - Q(s,a)]} \\
  & \added{\delta \gets R(s,a) + \gamma Q(s', a') - Q(s,a)} \\
  & \added{Q(s,a) \gets Q(s,a) + \alpha \delta}
  \end{aligned}
\end{equation}

\added{
In RL, the goal is to find the action to take given a state, which is the rough definition of the policy. Formally, a policy is a probability distribution that maps actions over given states. State-action-reward-state-action (Sarsa) is an on-policy model-free temporal difference learning algorithm which is shown in (\ref{eq:sarsa}). $Q(s,a)$ defines the Q-function which represents the expected total reward of taking action $a$ in state $s$. $R(s,a)$ represents the immediate reward of taking action $a \in A$ in state $s \in S$, $\alpha$ is the learning rate, and $\delta$ is the temporal difference error. For each episode, this equation is iterated starting from an initial state until a certain criterion is met. On-policy methods update the current estimate by the action selected from the policy. $Q(s,a)$ is updated using the $Q(s', a')$ where $a' \in A$ is selected from the current policy. On the other hand, off-policy methods such as Q-Learning update the current estimate by the action selected from another policy. Lastly, model-free methods do not require knowing the dynamics ($T$) of an environment.

The temporal difference learning is a \textit{bootstrapping} method where the previous estimate is used to update the new estimate. In Sarsa, only the current $Q(s,a)$ is updated, but Sarsa($\lambda$) uses eligibility traces. In eligilibity traces, every state visited during an episode is marked as eligible for update, and each iteration also updates the states that are marked as eligible. The eligibility of a state is decayed by $\lambda$ and $\gamma$.

The Q-function can be represented using a table of state and action pairs (tabular), or by a function approximator such as a neural network. Last but not least, there is the dilemma of exploration/exploitation. Exploration is gathering more knowledge and exploitation is choosing the best action with the current knowledge. The agent's objective is to maximize the total expected reward; therefore, it has to balance exploration/exploitation.
}

\deleted{
It is important to note that, Q-Learning is an off-policy method. Off-policy methods update the Q-value $Q(s,a)$ by assuming it will take the greedy action at state $s'$, which is independent of its current policy. State-Action-Reward-State-Action (Sarsa) is an on-policy method, which updates the Q-value by following the same policy. Last but not least, Q-function can be modeled using a tabular approach or a function approximator. Tabular methods are simple but spacious whereas approximator methods are complex but confined.
}

%Q Value is a measure of the overall expected reward assuming the Agent is in state s and performs action a, and then continues playing until the end of the episode following some policy π.

% \begin{equation} \label{eq:sarsa}
%   \added{ Q(s,a) \gets Q(s,a) + \alpha [R(s,a) + \gamma Q(s', a') - Q(s,a)] }
% \end{equation}

\subsection{Monte Carlo Tree Search}

Monte Carlo Tree Search (MCTS) \cite{Browne:2012} is a search method that iteratively expands the \added{search} tree in the preferred direction.\deleted{ This type of exploration results in an asymmetric tree where irrelevant regions are neglected.} MCTS \replaced{executes}{achieves this by executing} four consecutive steps iteratively until a certain condition occurs. This condition can be ---including but not limited to--- reaching the desired terminal node or expiration of the allowed computational budget.
These \added{four }steps are selection, expansion, simulation, and backpropagation. \added{The s}\deleted{S}election phase selects a node from the tree according to a Tree Policy. An acclaimed approach is to use the UCB1 \deleted{\added{(\ref{eq:2}) }}algorithm.

\deleted{
\begin{equation} \label{eq:2} 
UCB1 = \tilde{X_i} + 2C_p \sqrt{\frac{2 \ln \replaced{v}{n}}{\replaced{v}{n}_i}}
\end{equation}
}

\deleted{
$\tilde{X_i}$ is the average reward obtained from the $i^{th}$ child, $C_p$ is the exploration constant denoting how much we\deleted{ do} value exploration over exploitation. \replaced{$v$}{$n$} is the total number of times that the root is visited and \replaced{$v_i$}{$n_i$} is the total amount that the  $i^{th}$ child is visited. }This approach balances the exploration and exploitation of the search. In the expansion phase, one of the unexplored children of the selected node is added to the search tree. The simulation phase starts with this node, and a default policy is used to sample moves. The score obtained at the end of this simulation is used to update the values of the nodes, \replaced{starting from the simulated node propagating up to the root node}{from the simulated node to the root node}. This is the backpropagation phase. These four phases are repeated in this order until the computational budget expires. Afterwards\added{,} depending on a criterion\deleted{ \cite{Browne:2012}}, a child of the root node is returned.

\subsection{GVG-AI}

GVG-AI \cite{GVGAI:2019} is a framework that contains single/multi-player \replaced{2D}{two-dimensional} games. There are more than 120 games just for the single player, which include well-known games such as Mario, Zelda, and Sokoban. Due to the variety of games, GVG-AI poses a challenging and interesting environment. GVG-AI games also hold another special property\replaced{:}{,} they are all defined using a language called VGDL \cite{VGDL:2014}. This language defines the game rules in a specific game level, such as what will happen if the avatar attacks an enemy, or \deleted{when the avatar }interacts with a key.\added{ We slightly modified the GVG-AI to access all of the interactions amongst all sprites including the hidden sprites.}

\added{In this study, we consider the bugs between the game implementation and the game design. GVG-AI framework creates a game by transforming the game source code written in VGDL. Note that, we are not testing the internal engine that makes this transformation, but the created game. Thus, VGDL is the implementation, and the game scenario graph along with the game constraints are the game design.}

\section{Related Research}\label{sec:related_research}

\textbf{Game testing:} Software testing is a dynamic investigation for validating the software quality attributes. In the case of game software testing, these \deleted{quality }attributes \replaced{include}{have a quite wide range involving} cross-platform operability, aesthetics, performance in terms of time and memory, consistency and functional correctness in a multi-user environment \cite{Santos:2018}.\deleted{ Nantes et al. \cite{Nantes:2008} presented a semi-automatic general framework for game testing that combines artificial intelligence and computer vision. Their prototype was able to distinguish shadow map aliasing problems. Cho et al. \cite{Cho:2010} viewed the game as a black-box and instrumented tests based on specific scenarios. Ostrowski and Aroudj \cite{Ostrowski:2013} proposed a record and replay mechanism for Anno-2070 video game. Iftikhar et al. \cite{Iftikhar:2015} utilized UML class diagrams and state machines of the game to generate test sequences. Bécares et al. \cite{Becares:2016} applied Petri Nets to model a game. Petri Nets allowed them to execute high-level actions. An AI performed these high-level actions, and then the log generated by these tests were examined for bugs. Pfau et al. \cite{Pfau:2017} used discrete reinforcement learning to explore the states of an adventure game. Structures such as short-term and long-term memory determined the reward vector. Cicero framework \cite{Machado:2018} helped human testers to distinguish invincible barriers and fake walls in GVG-AI games. Loubos \cite{Loubos:2018} developed a framework that uses an RL agent to test the crafting system in Minecraft. However, most of these approaches do not consider an intelligent tester agent and even if it does, it only tests the given scenario without any modifications and lacks an overall coverage definition. Most importantly, human-data are only used to accomplish regression testing.}\added{ To validate these attributes, researchers proposed various methods to generate test sequences: record/replay \cite{Ostrowski:2013}, handcrafted scenarios \cite{Cho:2010}, UML and state machines \cite{Iftikhar:2015}, Petri nets \cite{Becares:2016}, and RL \cite{Pfau:2017}\cite{Loubos:2018}. However, when game environment changes, test sequences and scenarios such as \cite{Ostrowski:2013} \cite{Cho:2010} become obsolete and a manual tester effort is required to create a new test sequence. As opposed to these manual techniques, UML based techniques automate the process of test generation. However, generating sequences from UML runs into state explosion for larger games; and without a gameplay AI, it relies on the generated states to play the game. Hence, researchers employed AI to test games. In \cite{Becares:2016}, an AI played according to a petri net description of the game that contains high level actions, but they only generated test sequences that cover the game scenario. RL used in \cite{Pfau:2017} with short and long term memory is useful, but the approach is limited to the Point-and-Click games. The agent in \cite{Loubos:2018} can roam, but only the crafting system is tested using RL. More generalized approaches such as \cite{Machado:2018} experimented on two GVG-AI games for finding bugs. As the authors of \cite{Machado:2018}  stated, since the agent was a gameplaying agent, the agent was not required to find all of the bugs. Lastly, \cite{Romero:2018} introduced a team of agents with different purposes to test games. Nonetheless, all of these studies lack controlled experimentation.}
% Lovreto et al. \cite{Lovreto:2018} investigated and tested the games on the Google Play Store using Appium and the OpenCV library.
% An overview of the game testing methods and the overall process is, described by Redavid and Adil \cite{Redavid:2011}.

\textbf{Game Playing:} Researchers \deleted{have }applied RL and MCTS to numerous games, and there \replaced{are}{is} plentiful \replaced{studies}{research} on these topics.\deleted{ Tziortziotis et al. \cite{Tziortziotis:2014} used reinforcement learning with eligibility traces and temporal difference learning to play Ms. Pac Man game. Kormelink et al. \cite{Kormelink:2018} investigated the effects of different exploration methods in the Bomberman game, in their experiments, Max-Boltzmann exploration ranked first. Glavin and Madden \cite{Glavin:2015}, and Tastan and Sukthankar \cite{Tastan:2011} utilized reinforcement learning to train a video game bot in Unreal Tournament. Wang et al. \cite{Wang:2018} studied Q-learning on GGP games. They examined the effects of using MCTS in Q-learning to speed up convergence. In recent years, due to the success of deep learning, its applications in reinforcement learning are investigated. Mnih et al. \cite{Mnih:2015} merged reinforcement learning with deep neural networks to create a novel approach called deep Q-network. This network achieved surpassed human performance in several arcade games. More to this, Silver et al. \cite{Silver:2016} built a pipeline of various machine learning algorithms. Their program, AlphaGo, defeated the European Go champion, then the World Champion. Frydenberg et al. \cite{Frydenberg:2015} investigated the effects of four different MCTS modifications on GVG-AI games. Nelson \cite{Nelson:2016} examined the potential performance of the default MCTS controller in GVG-AI by altering the computational budget space. Horn et al. \cite{Horn:2016} calculated the difficulties of different features for AI agents. Bontrager et al. \cite{Bontrager:2016} evaluated the performance of several AI agents in GVG-AI framework. Childs et al. \cite{Childs:2008} considered transpositions and move groups in MCTS. In addition to these methods, Hierarchical temporal memory (HTM) provides a mechanism for reverse engineering human cortex. Building on this theory, Sungur and Surer \cite{Sungur:2016} investigated the Cortical Learning Algorithm (CLA) inside a 3D virtual world. Although the aim of these papers was to create better agents in game playing, our purpose is to create an agent that tests the game by the playing with respect to test goals.}\added{ For example, Sarsa($\lambda$) is used as a game playing agent in Ms. Pac Man \cite{Tziortziotis:2014} and to create a human-like agent in Unreal Tournament \cite{Glavin:2015}. Although the aim of these papers was to create better agents in game playing, our purpose is to create an agent that tests the game by playing with respect to test goals. In general game playing, MCTS's aheuristic and anytime characteristics are prevalent. In order to increase the performance of vanilla MCTS, researchers proposed several modifications \cite{Frydenberg:2015} \cite{Soemers:2016}. Amongst them, knowledge-based evaluation (KBE) \cite{Perez:2014} is found beneficial. Therefore, we used KBE enhancement in our MCTS.
% In recent years, due to the success of deep learning, its applications in reinforcement learning are investigated. Mnih et al. \cite{Mnih:2015} merged reinforcement learning with deep neural networks to create a novel approach called deep Q-network. This network achieved surpassed human performance in several arcade games.
}

\textbf{Learning From Humans:} Incorporating domain knowledge is prominent, both in RL and MCTS, yet defining the correct set of rewards is a hurdle. Moreover, even \deleted{us,} humans learn better when \replaced{guided by or imitating an expert}{guided by an expert or when imitating an expert}. Inverse reinforcement learning (IRL) is the study of extracting a reward function, given an environment and observed behavior, which is sampled from an optimal policy \cite{Ng:2000}.\deleted{ Abbeel and Ng demonstrated two methods for extracting this behavior \cite{Abbeel:2004}. Maximum likelihood with gradient optimization is utilized in \cite{Ziebart:2008}\cite{Babes-Vroman:2011}. Wulfmeier et al. \cite{Wulfmeier:2015} extended Maximum Entropy IRL using deep architectures that can extract non-linear reward functions. These IRL methods assumed a single near-optimal policy. On the other hand, Michini and How \cite{Michini:2012} used Bayesian nonparametric mixture model to automatically partition the trajectory and discover sub-goals, {\v{S}}o{\v{s}}i{\'c} el al. \cite{Sosic:2018} generalized this model to unseen states using a distance metric. Rhinehart and Kitani \cite{Rhinehart:2018} exploited stop detection to segment the trajectory and capture goals. Consequently, many researchers used IRL for various goals. Tastan and Sukthankar \cite{Tastan:2011} harnessed IRL to extract weights from collected human trajectories. Ivanovo et al. \cite{Ivanovo:2015} investigated IRL to extract weights from AI opponents and used these weights to model them. Nevertheless, there are other methods to extract knowledge from the collected data. Ortega et al. \cite{Ortega:2013} employed artificial neural networks to learn from recorded data. Dobre and Lascarides \cite{Dobre:2015} utilized function approximation to capture state values. Silver et al. \cite{Silver:2016} applied supervised learning to train a network that proposes a move given a game state. Khalifa et al. \cite{Khalifa:2016} modified the selection step of MCTS to mimic human playing at GVG-AI corpus. Devlin et al. \cite{Devlin:2016} determined the weights of different actions using cross-entropy. These weights are used to guide an MCTS agent's actions. Gudmundsson et al. \cite{Gudmundsson:2018} applied supervised learning to train a deep neural network and utilized this network to predict the most human-like action in the Candy Crush game. All of the examples presented in this subsection use the recorded data to create an agent that can apply human knowledge. Since our data are based on human tester trajectories and considering the whole data may not be optimal \cite{Michini:2012}, these trajectories should be split to achieve locally optimal trajectories \cite{Sosic:2018}. Nonetheless, the approaches that promoted splitting are exercised on the same domain. Whereas we want to transfer this knowledge to other levels. Moreover, the applications of human data were not used to create a tester agent.}\added{ IRL methods exist in order to extract reward function when i) trajectories that are sampled from the same policy \cite{Abbeel:2004} \cite{Ziebart:2008} \cite{Wulfmeier:2015}, ii) trajectories that are sampled from different policies \cite{Babes-Vroman:2011}, and iii) trajectories that are better explained with multiple sub-goals \cite{Michini:2012} \cite{Sosic:2018}, are given. The trajectories that were collected from testers fit into the last category since the testers can test several goals in the same run. However, as noted by \cite{Sosic:2018}, \cite{Michini:2012} fails to generalize to unseen states and the approach in \cite{Sosic:2018} finds sub-goals in the same level. Therefore, we propose MGP-IRL overcome these drawbacks. Additionally, human-like agents are used in games such as Unreal Tournament \cite{Tastan:2011}, Super Mario \cite{Ortega:2013}, Catan \cite{Dobre:2015}, GVG-AI \cite{Khalifa:2016}, Spades \cite{Devlin:2016}, and Candy Crush \cite{Gudmundsson:2018}. However, we focus on human-like tester agents.}

\section{Methodology}\label{sec:methodology}

We propose \replaced{an agent based system}{to generate agents} to automate game testing to detect the discrepancies between game design and game \replaced{execution}{implementation}. To this purpose, two types of \replaced{test goals}{agents} are created: a) synthetic \replaced{test goals}{agents} based on the game scenario, rewarding all valid and some invalid transitions, \added{and} b) human-like \replaced{test goals}{agents} that are trained from human tester data. Using these \added{test goals in RL and MCTS} agents, we generate test sequences and check if the game software behaves according to the game \replaced{constraints}{rules} automatically.

\added{In this study, the term ``interaction'' does not refer to the interaction definition used in GVG-AI framework, but it refers to our definition of interaction, which is proposed for game testing purposes.}

In the following sections, first, we define the interaction state. Then, we present our approach for synthetic \replaced{test goal}{agent} creation. We continue by explaining our method in learning test \replaced{goals}{behavior} from the human data\deleted{ for intelligent agents}. Finally, we describe how to generate test sequences\replaced{, and our proposed oracle}{ based on these two types of agents and detect the defects}.

\subsection{Interaction State}\label{sec:methodology:is}

Testers exercise a game by following several strategies \cite{Redavid:2011} which lead them to interact with various aspects of the game. Testers mark these interactions as ``tested'' in a memory. This memory prevents testers from executing the same interactions. Memory can be a pen and paper, a tool or the intangible memory of the testers.

Interactions that progress the game, such as picking up a key can be modeled using an MDP. On the other hand, it is difficult to model the interactions that do not advance the game, such as trying to pass through a wall. If the game does not allow this behavior, the player should stay in the same position. Moreover, if a positive reward is obtained with this interaction, an agent can repeat \replaced{the hitting the wall}{this} interaction infinitely. \deleted{The main problem is that the states ---before and after the interaction--- are the same.} A memory can solve this \replaced{state representation}{problem} by marking this wall as ``tested''. The game state does not \deleted{progress from}\added{record} this interaction, but the memory \replaced{does}{is updated}. Therefore, the MDP formulation becomes simpler\added{ when we use both game state and interaction state together}.

An idea of memory is used by Pfau et al. \cite{Pfau:2017} to explore the states of \deleted{an}\added{a Point-and-Click} adventure game. \added{They used this memory idea to adjust the reward of each available action. However, we use memory to record the interactions performed during testing. }\deleted{ However, i}\added{I}n this study, we examine automated testing approaches on \added{2D }grid games. Therefore, we use a grid\replaced{-based memory model}{ to model the memory}. This grid is referred to as the interaction state, as it marks interactions. \added{The i}\deleted{I}nteraction state is a supplementary state to the game state. \added{The g}\deleted{G}ame state is a grid that holds the positions of sprites in the game. Using only the game state to model testing behavior by MDP is inadequate for the following reasons. First, only specified \added{VGDL }game rules can alter the game state; thus, some of the interactions will not alter the game state at all. Second, due to a bug, even state changing interactions may not manifest on the game state.

In \added{2D }grid games, interactions occur between sprites. Therefore, we \replaced{propose}{define} an interaction as a tuple: $\zeta = {<}\eta_0,\eta_1,Pos,Dir,Type, Avatar_{State}{>}$ where $\eta_0, \eta_1 \in \Gamma$ and $\Gamma$ is the set of sprites, $Pos: {<}x,y{>}$ is the position \deleted{of}\added{where the} interaction\added{ took place}, $Dir \in \{\uparrow, \downarrow, \leftarrow, \rightarrow\}$ is the direction of the interaction\added{ calculated from the first sprite's direction ($\eta_0$)}, $Type \in \{Move,\deleted{Attack}\added{Use}\}$\added{ where $Use$ can be mapped to actions such as \textit{Attack}}, $Avatar_{State}$ represents the states that an avatar can take\added{ ---for VGDL they are listed under avatar definition}. \added{We should note that, the first three parameters are mandatory while the rest are optional.}
% These last three parameters can be dropped depending on the tester.

We use a grid-base\added{d} model to represent the interaction state since we focus on grid games.\deleted{We chose grid approach as opposed to a set to model the interaction state since fast calculation of state's hash is advantageous in tabular RL methods.} An interaction state is modeled using a 3D grid \replaced{to identify}{to separate} the following concerns: the tester may prioritize testing a sprite from all directions, the tester may exercise an interaction more than once, \added{and }the tester may choose to differentiate between a movement and \deleted{an attack}\added{\textit{Use}}. A 3D grid solves these concerns while keeping a single number in every grid cell. This number counts how many times \replaced{an}{this} interaction \added{has} occurred. Every layer in this 3D grid is a 2D grid\deleted{,} which has the width and height of the game grid. In the final implementation, we used 12 2D grids. First four layers represent the movement interactions done between the \replaced{a}{A}vatar and \deleted{the}\added{other} sprites. Each of these four layers represents a different direction. Next four layers represent interactions with \deleted{an attack}\added{\textit{Use}}, again one for each direction. The last four layers represent the interactions \deleted{done from other moving types}\added{that do not fit the aforementioned levels}; such as when the avatar pushes the water bucket\replaced{ and}{,} the water bucket also moves or when an enemy interacts with another sprite. Once more, it holds a layer for each direction.\deleted{ We considered that the tester might not differentiate between directions; then, the same cell positions in all of the four layers are incremented.}

\added{Although we proposed an interaction state for grid games, it can be applied to other games as well. The basic idea is to duplicate the visual environment for the purpose of marking tested positions, and this duplication process can be automated. Furthermore, additional layers can be added depending on what needs to be differentiated between different categories: movement, use, and so on. If the objects such as keys or doors have more importance, rather than creating a replica of the environment, an array containing just these preferred objects can be created. 
% Lastly, if we put an upper limit on the repetition limit, we can pack the layers into more efficient structures.
}

For the rest of this paper, we use the following definitions. A Game is a tuple ${<}\mathcal{G}, \mathcal{I}, A, \Gamma, \delta, ||\zeta||{>}$, where $\mathcal{G}$ is the set of game states, $\mathcal{I}$ is the set of interaction\deleted{s} states, $A\added{=\{\uparrow, \downarrow, \leftarrow, \rightarrow, Use, Nil\}}$ is the set of all actions\deleted{,}\added{ where the availability of $Use$ depends on the game, and $Nil$ signifies taking no action.} $\Gamma$ is the set of sprites and $\delta$ is a transition function that takes an action and outputs the next game state, interaction state, \deleted{and} the interactions\deleted{.}\added{, and $||\zeta||$ is the set of all interactions.} \deleted{Therefore, }$S$ in MDP \added{(see Section \ref{sec:pre:rl}) }is defined as $S: \mathcal{G} \times \mathcal{I}$\deleted{ and $s: {<}g, i{>}$ where $s \in S, g \in \mathcal{G}, i \in \mathcal{I}$}.
%, $\delta$ yields a new game state by applying the chosen action to a game state, $\delta(s, i, a){=} s', i', \zeta_{0..n}: s, s' \in \mathcal{S}, i, i' \in \mathcal{I}, a \in \mathcal{A}$.

Lastly, a feature is a tuple $\phi{=}{<}\eta_0, \eta_1, Weight, Method,\\ Type, Rep, Avatar_{State}{>}$ where $\eta \in \Gamma$, \added{\textit{Weight} is the reward obtained from this feature.} \textit{Method} represents the direction preference of the tester. We have two different proposals for this parameter $\{Each,All\}$. \textit{Each} is used to differentiate different directions\deleted{,} whereas \textit{All} is used for considering all of the directions to be the same. \deleted{\textit{Weight} is the reward obtained from this feature.} \added{\textit{Type} is \{Move, Use\}.} \textit{Rep} limits the number of times that the reward can be obtained. \added{Lastly, \textit{Avatar\textsubscript{State}} is the same as that of interaction.}

\added{An action taken on a game state generates interactions, and interactions are matched with features which are used to calculate the reward. When an interaction $i$ occurs, the feature $\phi_i$ corresponding to this interaction is retrieved from the feature set of the game using $\eta_0, \eta_1, Method, Avatar_{State}$ values of the interaction. Then, the \textit{Rep} of $\phi_i$ is compared with the number of times the interaction $i$ has occurred, which is stored in the interaction state.} \deleted{When an interaction occurs, its value ---stored in the interaction state--- is fetched, and this value is compared against the \textit{Rep} parameter in the feature.} If this value is less than or equal to the \textit{Rep}, the reward defined in \textit{Weight} is acquired and the value in the cell is incremented.\added{ \textit{Each} only updates a single cell, whereas \textit{All} updates all of the cells.} \added{This is how the \textit{reward function} is calculated.} \replaced{Lastly}{Therefore}, if an interaction does not match with a feature, the interaction state will stay the same.

\begin{figure}
  \begin{footnotesize}
  \begin{Verbatim}[commandchars=\\\{\}]
  SpriteSet
    floor > Immovable img=oryx/floor3 \deleted{hidden=True}
    goal  > Door img=oryx/doorclosed1
    key   > Immovable img=oryx/key2
    sword > OrientedFlicker img=oryx/slash1
    movable >
      avatar  > ShootAvatar stype=sword
        nokey   > img=oryx/necromancer1
        withkey > img=oryx/necromancerkey1
    wall > Immovable img=oryx/wall3
  InteractionSet
    movable wall > stepBack
    nokey goal   > stepBack
    goal withkey > killSprite
    nokey key    > transformTo stype=withkey
                     killSecond=True

  \end{Verbatim}
  \end{footnotesize}
  \caption{Simplified SpriteSet and InteractionSet in VGDL}
  \label{fig:vgdl}
\end{figure}

\begin{figure}[]
  \centering
  \subcaptionbox{Game State\label{fig:game_interaction_a}}
      {\includegraphics[width=0.48\columnwidth]{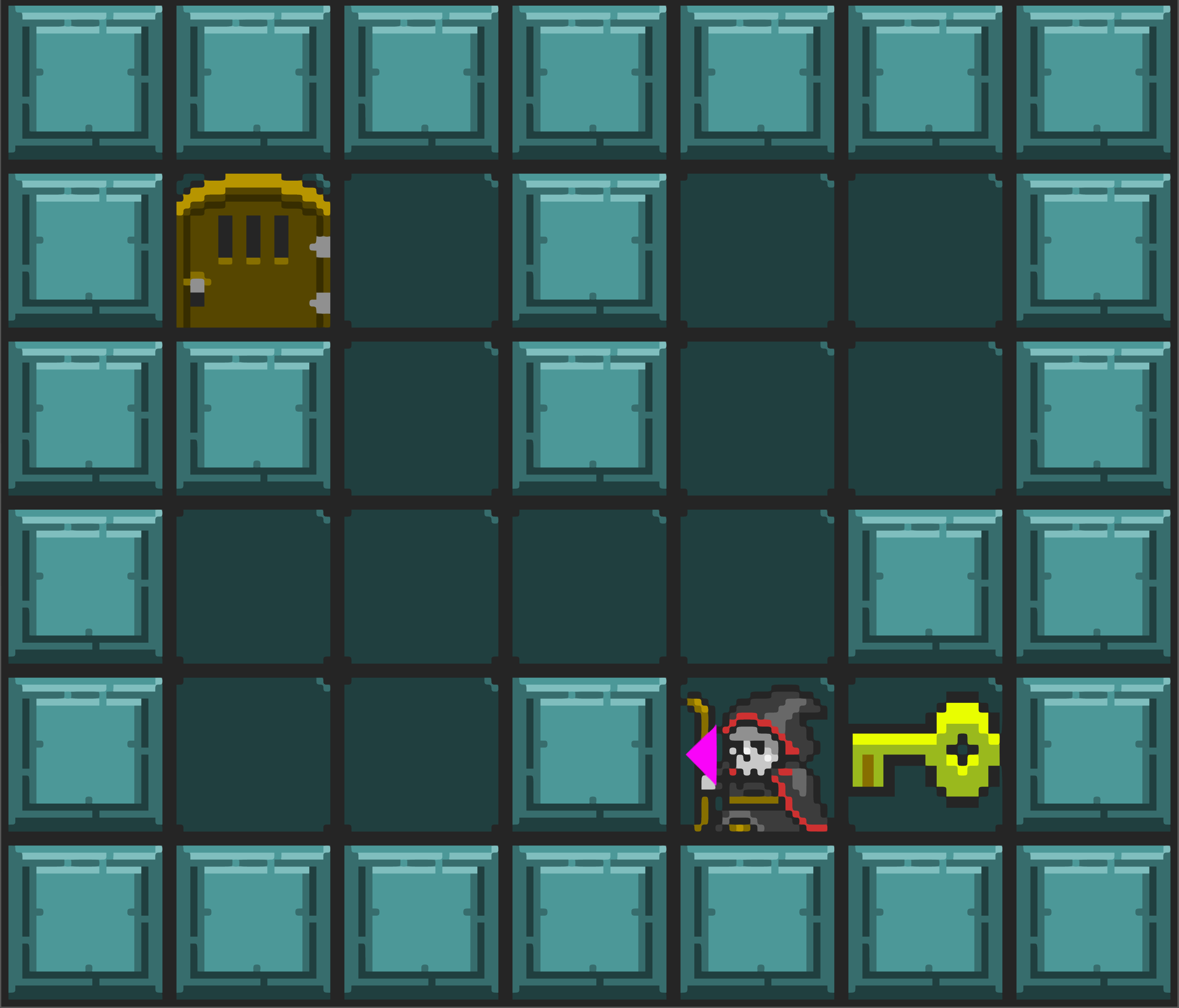}} \hfill
  \subcaptionbox{Interaction State\added{ with Each as Method}\label{fig:game_interaction_b}}
      {\includegraphics[width=0.48\columnwidth]{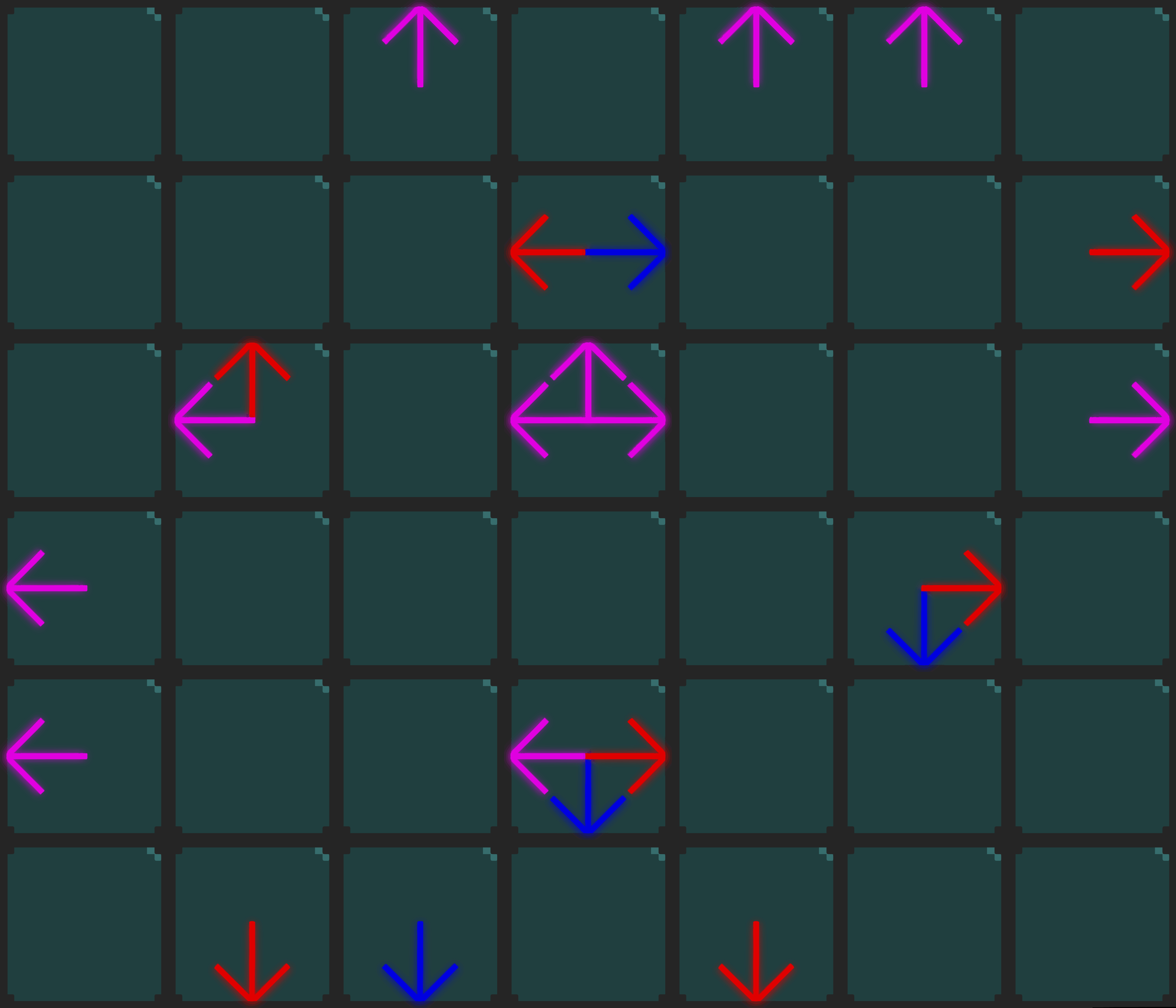}} \hfill
  \subcaptionbox{Interaction State\added{ with All as Method}\label{fig:game_interaction_c}}
      {\includegraphics[width=0.48\columnwidth]{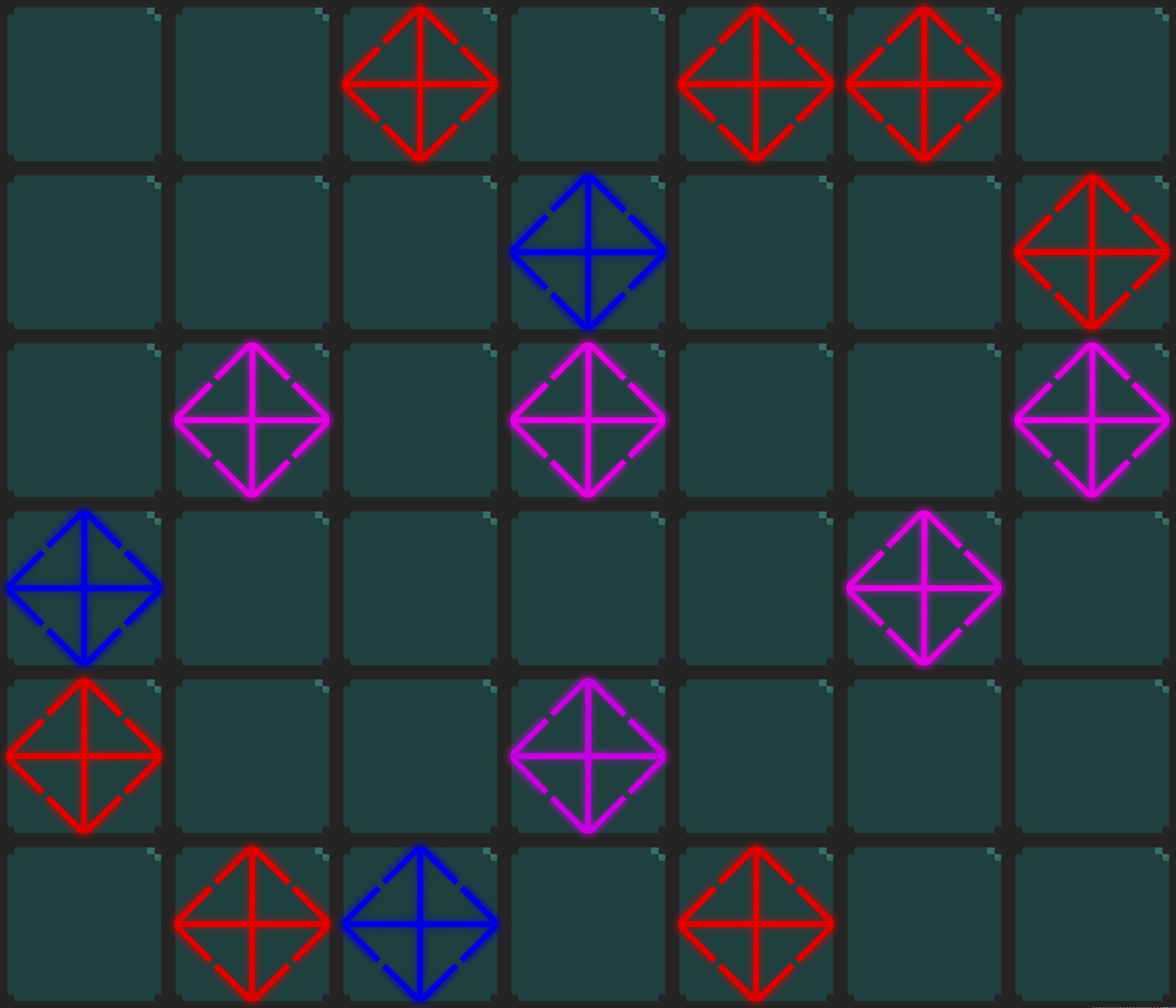}} \hfill
  \subcaptionbox{Interaction State Layers\added{ from \figurename{ \ref{fig:game_interaction_b}}}\label{fig:game_interaction_d}}
      {\includegraphics[width=0.48\columnwidth]{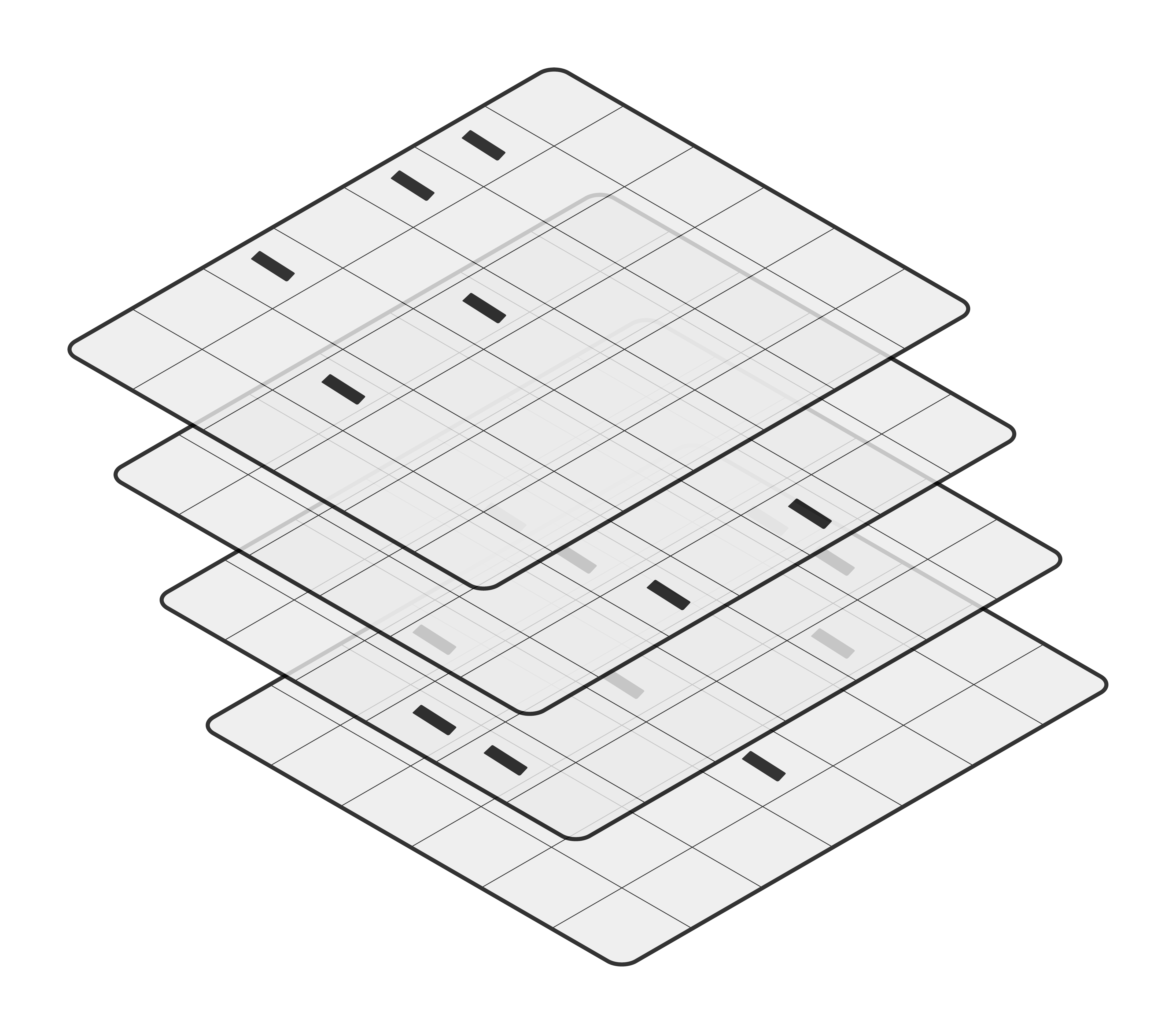}} \hfill
  \caption{Game State and Interaction State}
  \label{fig:game_interaction}
\end{figure}

To illustrate the game and \added{the }interaction states, consider the game \replaced{transcribed}{described} in VGDL in Fig. \ref{fig:vgdl}. \replaced{W}{In this running example, w}e illustrate the game state and two different interaction states in Fig. \ref{fig:game_interaction}. A game state that contains the sprites $\Gamma {=} \{Avatar, Door, Floor, Key, Wall\}$ is shown in Fig. \ref{fig:game_interaction_a}. The magenta triangle on the \replaced{a}{A}vatar represents its direction\added{, and $Avatar_{State}{=}NoKey$}. In Fig. \ref{fig:game_interaction_b} and Fig. \ref{fig:game_interaction_c}\added{,} two interaction states are shown, where interactions in the first four layers are represented with red color\replaced{ while}{,} interactions in layers four to eight are represented with blue color. Their combination is depicted in magenta. The arrows represent the direction of \added{the} interaction saved in the state. In Fig. \ref{fig:game_interaction_b}, since individual arrows are shown, it can be understood that the features' \textit{Method} parameter is \textit{Each}, whereas in Fig. \ref{fig:game_interaction_c} individual arrows are not shown, thus the features' \textit{Method} parameter is \textit{All}. In Fig. \ref{fig:game_interaction_d} the first four layers of Fig. \ref{fig:game_interaction_b} \replaced{are}{is} dissected, \added{and }for convenience \added{cells that have }zeros are left empty. Recall that, the numbers in cells show how many times an interaction is performed. 

In this grid, the top-left corner is ${<}0,0{>}$. The avatar in Fig. \ref{fig:game_interaction_a} is at position ${<}4,4{>}$.\deleted{ The direction is shown with arrows in the grids.} For example, in Fig. \ref{fig:game_interaction_b}, the direction of the interaction at position ${<}2,0{>}$ is $\uparrow$. There could be more than one interaction at a position. For example, in Fig. \ref{fig:game_interaction_b} at position ${<}3,4{>}$ all four directions except $\uparrow$ are exercised at this cell.\deleted{ This cell also shows that the direction preference of the tester, i.e. the value of \textit{Method}, is \textit{Each}.} In Fig. \ref{fig:game_interaction_b}, we see that the agent executed ${<}Avatar,Wall,{<}2,0{>},Move,\uparrow, NoKey{>}$ and the feature was ${<}Avatar,Wall,1,Move,Each,1,NoKey{>}$ since there is only an upward arrow. In Fig. \ref{fig:game_interaction_c}, the following features are used  ${<}Avatar,Wall,1,Move,All,1, NoKey{>}$ (see the blue and magenta marked cells), ${<}Avatar,Wall,1,\deleted{Attack}\added{Use},All,1, NoKey{>}$, (see the red and magenta marked cells). This interaction state shows that, in the top middle walls, the tester executed the \deleted{attack}\added{\textit{Use}} and \replaced{\textit{Move}}{move} interactions.

\subsection{\replaced{Test Goal Generation}{Tester Agents}}
\subsubsection{Synthetic \deleted{Agents}\added{Test Goals}}

Automation of the testing steps is crucial as it speeds up development while reducing the testing effort. Although approaches like playing a pre-defined or pre-captured scenario are effective, they are no longer usable when the game \replaced{layout}{design} changes. 
In this study, we present \deleted{an approach based on the game scenario graph. We call the tester agents created using this method as synthetic agents since no human data are used.}\added{a game scenario graph-based approach to create synthetic test goals.}

\added{We use the game scenario graph to generate various scenario paths using a graph coverage criterion.}\deleted{For a synthetic agent, paths to be covered are generated using a game scenario graph and a coverage criterion.} In our case, the game scenario graph defines only allowed transitions. These\added{ generated} paths target different game routes in the scenario.\deleted{ and agents can play these paths to check if the game implements the scenario.}\added{ Our aim is to create test goals from these paths that an agent understands. In this regard, we use features to guide the MCTS or RL agent towards these generated test goals. Hence, our agents can verify if the game scenario is implemented correctly in a game.
}

In addition to checking if the game implements the scenario, we also want to check whether the game implements some additional behavior that \deleted{are}\added{is} \textbf{not} in the scenario, as any tester would do. This promotes the agent to ask questions like: What happens if I attack the key? Can I pass through \added{the }walls? We generate a list using all combinations \deleted{\cite{Kuhn:2015}} of the following four parameters of a feature ${<}\eta_0,\eta_1,Type, Avatar_{State}{>}$ and this list is referred to as modifications list. For games with high sprite count, pair-wise combinatorial \cite{Kuhn:2015} strategy can be used in the generation of the modification list. We initially prune the modification list by restricting $\eta_0$ to be movable.

\deleted{Synthetic agent generation algorithm is as follows: First, given a game scenario graph and a coverage criterion, paths to be covered are generated. Second, for each path in the form $[n_1,n_2,...,n_M]$, we use the edges $[(n_1,n_2),...,(n_{M-1},n_M)]$ to convert this path into a sequence of features. For example, if edge transition occurs by picking up the key, a feature, which will allow the synthetic agent to perform that action, is generated.}\added{Synthetic test goal generation algorithm is as follows: The developer gives a game scenario graph, such as \figurename{} \ref{fig:scn_graph}, and a table that maps the edges to abstract features. A feature is abstract when \textit{Weight}, \textit{Method}, and \textit{Rep} parameters are empty. Given this scenario graph and a coverage criterion, the system first generates paths that need to be covered during test execution. Then, each path of the form $[n_1,n_2,...,n_M]$ is converted into a sequence of features by replacing each pair of nodes $(n_i,n_j)$, i.e. edge, with the abstract feature retrieved from the given table. After this conversion, the paths are discarded, and the process continues with the feature sequences. }Next, \added{each of }this generated feature sequence is modified \added{by inserting an extra feature from}\deleted{using} the modification\deleted{s} list. \replaced{This extra feature}{These modifications} can be inserted to any place in the sequence. \replaced{W}{However, w}hen modifying a sequence of features, the Manhattan distance between the candidate feature from the modification list and every feature in the sequence should be at least one. \deleted{Manhattan distance is calculated using $\eta_0,\eta_1,Type, Avatar_{State}$ parameters of a feature.}\added{Manhattan distance is calculated by comparing the equality of $\eta_0,\eta_1,Type, Avatar_{State}$ parameters \replaced{of}{between} features. If a compared parameter is the same, its distance is zero and one if they are not the same.} \replaced{This control prioritizes the features that have not been included in the original feature sequence.}{This control prevents adding transitions that will be checked by the original path.}\deleted{ These modifications can be applied in numerous ways. }

For a \replaced{feature sequence}{path} that has $M$ \replaced{features}{nodes} and $K$ number of modifications, if we choose to insert all \added{modifications }to the \replaced{sequence at the same time}{same path}, we can insert up to $M \times K$ modifications ---not considering the order of modifications. Nevertheless, this approach is inadequate for testing, a bug found \replaced{while agents using a feature from modification list}{on the initial edge} might crash the game and the bugs \replaced{will not be noticed which can be found by using other features from this list}{that dwell in the second edge will not be found}. Therefore, to prevent this ``masking'' and to reduce the complexity of the \replaced{feature sequence}{path} to be played, \added{we copy the original feature sequence $M \times K$ times. Then insert one modification into each copy.}\deleted{for every \replaced{transition}{edge} and for every modification, we copy the original sequence of features and insert a single modification\deleted{ to the edge}.} \added{We also include the original feature sequence that is not modified. }Therefore, from a single \replaced{feature sequence}{path}, $M \times K \added{ + 1}$ number of feature sequences are generated at most. \deleted{We also include the original feature sequence in the final paths to be on the tested list.} It should be noted that\added{ an abstract feature is concretized by setting} the other three feature parameters \deleted{are determined }manually. In this paper, \textit{\replaced{Rep}{Repetition}} parameter is set depending on $\eta_1$: \deleted{three}\added{3} if it is a movable sprite, \deleted{one}\added{1} if it exists in abundance such as walls, else set as \deleted{two}\added{2}. \textit{Method} is set as \textit{All}, and the parameter \textit{Weight} is set as one.

This approach generates features that will guide the agent through different parts of the game. For a more complex graph, these features can become difficult to play. Since every edge in \replaced{a game scenario}{the} path corresponds to a feature, the number of features \deleted{an agent has to play} scale\deleted{s} with the number of edges in \replaced{this}{the} path. Hence, in this study, we divide the overall path, influenced by the sequential approach of Rhinehart and Kitani \cite{Rhinehart:2018}. This division is crucial due to two factors: first, we want the agent to execute the feature sequence in the intended order so that the agent traverses the scenario graph in the intended order; second, due to a bug the overall path may not be played, but separation helps pinpoint exactly in which part the problem occurred.

We present a different goal state definition from \cite{Rhinehart:2018}\cite{Michini:2012}\cite{Sosic:2018}. Our goal state \added{---test goal---} is defined as $S_h$ where $S_h \subseteq S$. In  \added{the} case of a goal such as testing all walls or covering all empty spaces, there are many states in $S_h$ that represent this goal. Moreover, we should be able to define a goal where the agent tests a single wall. However, this definition does not allow a flexible mechanism to specify this goal. To this purpose, \replaced{we propose a criterion for each feature. A criterion specifies the percentage of $\eta_1$ to be tested}{we propose criteria which include definitions such as wall percentage to be tested and the percentage of space to be explored. }\replaced{}{We define a criterion for each feature}, and the combination of features and criteria constitutes a goal. Consequently, while feature\added{s} guide\deleted{s} the agent through the grid, the criteria check whether the agent fulfilled the goal. 

When the goal is fulfilled, i.e. the agent is in $S_h$, the agent moves on to the next goal in the sequence. 
\added{
The agent may not fulfill a goal due to two reasons. First, the goal may be infeasible, such as attacking a door which is hidden. Second, a bug may prevent reaching a goal. Hence, we may choose to terminate or move on to the next goal if an agent does not reach a goal. In our implementation, we made this choice optional. 
}
Furthermore, to assist exploration, the feature referred to as the exploration feature ${<}Avatar, Floor, All, Move, 0.01, 1, Avatar_{State}{>}$ is added to every goal, and $Avatar_{State}$ parameter is copied from the other feature in the same goal. 
\deleted{The goals in the synthetic agent only includes the bare necessary features, and if these features are far way in the game grid, this exploration feature helps the agent to wander around and interact with the required feature.
}
In synthetic \deleted{agent}\added{test goals}, we set criterion values to 100\% for the features from the original sequence and 0\% for the exploration feature. The agent acquires a positive reward for traveling\added{,} but exploration is not required for the agent to pass the criteria.
Formally, the goal $h$ consists of features and a criterion for each feature $\{(\phi_0,c_0),(\phi_1,c_1),...,(\phi_n,c_n)\}$, where \replaced{$c_0,...,c_n$ are positive real numbers}{$\{c \in \mathbb{Q}~|~c \geq 0 \}$}. Hence, \replaced{the synthetic test goal approach generates}{the result of a path is} a sequence of goals $\mathcal{H} {=} (h_0,...,h_n)$.
\\
% The goal idea in \cite{Rhinehart:2018} is adopted, and from each node, in the path, a goal is created, hence a sequence of goals represents a path, and a goal represents a test purpose. Nevertheless, defining transitions becomes problematic while defining goals for nodes e.g. how the agent will understand whether it has tested movement against every wall. Since there is no clear state definition such as in picking up the key, we introduce criteria. Criteria include definitions such as wall percentage to be tested, the percentage of space to be explored. Therefore, the combination of features and criteria constitutes a goal. Consequently, while feature guides the agent through the grid, the goal checks whether the agent has fulfilled the criteria. When the criteria are fulfilled, the agent moves on to the next goal. Hence, every modified path is converted to a sequence of goals. In synthetic agent, we set criterion values to 100\%.
\added{
% For the game depicted in Fig. \ref{fig:game_interaction_a}, we have a game scenario graph that consists of three nodes $\{{<}\neg Key,\neg Door{>},{<}Key,\neg Door{>},{<} Key, Door{>} \}$, and the path sampled from this graph will contain these three nodes.  We generate a modification list consisting of 16 modifications using the combinations of ${<}\eta_0,\eta_1,Type, Avatar_{State}{>}$. In our example, the first edge updates the $Key$ attribute.  We convert this change into a feature using a table which is designed by the game designer. This table holds a predetermined feature for each attribute. In this example, this edge is converted into a feature ${<}Avatar,Key,1,Move,All,2, NoKey{>}$.  The sampled path is converted into: $\{{<}Avatar,Key,1,Move,All,2, NoKey{>}, {<}Avatar,Key,\\1,Move,All,2, NoKey{>}\}$, and assume that we modify it using ${<}Avatar,Wall,1,Use,All,1, NoKey{>}$. We can insert this feature into three different places. Although, it is meaningless to add this after picking the key, a bug may prevent picking up the key. Hence, we have made this type of additional pruning optional. After inserting the modification, this path will have three features. Each of these features are converted into a goal, the feature part is copied and for their criterion 100\% is added. Hence we get three goals, which are linked in the original sequence order.
}

\subsubsection{Human-Like \deleted{Agents}\added{Test Goals}}

Beta-testing is an invaluable part of the game development process. \replaced{Human testers participating in beta-testing use their expertise and heuristics to uncover various bugs.}{Human testers participating in beta-testing exercise their expertise and the heuristics that the test experts apply become valuable. This ad hoc testing behavior \cite{Redavid:2011} can uncover various bugs.} During any test phase, the actions of each participant can be recorded as trajectories. A trajectory $\tau$ is a sequence of actions $[a_0,...,a_n]$ where $a_i \in A$ and $0 \leq i \leq n$. A trajectory itself does not represent anything meaningful, but, when the trajectory is replayed, it exhibits the \replaced{intentions}{intuitions} of a tester. Therefore, in \added{the }literature, the collected trajectories are used in regression testing \cite{Ostrowski:2013}.\deleted{ However, an update to the game requires testing to be repeated \cite{Becares:2016}.} In this study, we aim to capture the human testers' expertise and automate the test generation by learning from the actions of these testers\added{ instead of repeating/replaying them}. In this regard, \deleted{IRL}\added{inverse reinforcement learning (IRL)} is chosen to grasp this \replaced{expertise}{behavior}. IRL assumes that the trajectory is near\deleted{ly}-optimal, but during ad hoc testing this assumption may not hold. Moreover, the human tester may perform a complex sequence of actions that cannot be modeled by linear weights \cite{Michini:2012}\cite{Sosic:2018}. Therefore, we \replaced{propose to}{} automatically partition these trajectories \replaced{so}{in a way} that these partitions are near\deleted{ly}-optimal\added{, as described below}.

We propose MGP-IRL to capture tester expertise, which is presented in Algorithm \ref{alg:mgp_irl}. First, at line \ref{alg:mgp_irl:split}, the algorithm replays the actions in the trajectory and splits them to minimal trajectories and interactions. We split the trajectory at points where interactions change. We define change as any variation in $\eta_0 ,\eta_1, Type$ parameters of an interaction. At line \ref{alg:mgp_irl:init}, \added{the }set of previous features $\Phi_a$ is initialized as empty set, sequence of previous trajectory $\tau_a$ is set as empty sequence, the previous likelihood threshold $\kappa_a$ is initialized as zero, and the goal sequence $\mathcal{H}$ is set as empty sequence.

In lines \ref{alg:mgp_irl:create_feature} to \ref{alg:mgp_irl:combine_feature}, the current segment is converted to a feature $\phi_i$ and $\Phi_b$ becomes the union of the feature $\phi_i$ and the previous features $\Phi_a$. This feature discovery allows the algorithm to employ non-zero weight strategy for unobserved interactions \cite{Abbeel:2004}. This strategy shrinks the feature space and supports learning with fewer expert trajectories. Note that parameters such as $Weight, Method, Rep$ cannot be captured from interactions, these parameters are left empty at this step. At line \ref{alg:mgp_irl:analyze_rep}, the trajectory $\tau_b$ is replayed with the features $\Phi_b$ to find the repetition count of $\Phi_b$. Without this step, our IRL procedure will be ill-posed\deleted{,} since there is a reward that the agent can acquire for unknown amount of times. 

At line \ref{alg:mgp_irl:irl}, IRL is applied to the trajectory $\tau_b$ to find the weights of the features $\Phi_b$. Next, at line \ref{alg:mgp_irl:likelihood}, the likelihood \cite{Babes-Vroman:2011} of the trajectory $\kappa_b$ is calculated using $\Phi_b$. This likelihood estimate is used to decide whether we should combine trajectories and features. We know that trajectory $\tau_a$ can be sampled from a policy $\pi_a$ that uses $\Phi_a$ with the likelihood of $\kappa_a$\added{,} and question whether $\tau_i$ should be added to $\tau_a$ which creates a new policy $\pi_b$ that uses $\Phi_b$ with the likelihood of $\kappa_b$. \added{If the combination of these trajectories is more likely, the the difference of $\kappa_a$ - $\kappa_b$ is expected to be negative.}\deleted{Therefore, the difference of $\kappa_a$ - $\kappa_b$ is examined. If this difference is lower than zero, it means that \added{the }combination is more \deleted{meaningful}\added{likely to be executed by an agent under found weights. However, \added{the }larger this difference, the less likely \added{it is }to be executed by the agent}.} 

\added{If the condition at line \ref{alg:mgp_irl:check} holds, the previous features, likelihood, and trajectories are replaced with the current features, likelihood, and trajectories. If we cannot combine the trajectories, at line \ref{alg:mgp_irl:create_goal}, the previous segment is converted into a goal by calculating the criterion of each feature in $\Phi_a$ ---if the feature holds a non-negative reward.}

  During experiments, we \replaced{examine}{consider} the effect\deleted{s} of \replaced{the}{this} threshold \deleted{and this is controlled by }$\kappa_T$. \added{The lower the threshold, the more likely the agent can repeat these interactions. As the threshold increases, the algorithm behaves more similarly to the internal IRL algorithm.}
\deleted{If the condition at line \ref{alg:mgp_irl:check} holds, the previous features, likelihood, and trajectories are replaced with the current features, likelihood, and trajectories. If we cannot combine them, at line \ref{alg:mgp_irl:create_goal}, the previous segment is converted into a goal by calculating the criterion of each feature in $\Phi_a$ ---if the feature holds a non-negative reward.}

To calculate the criteria, first, the previous trajectory $\tau_a$ is replayed to count how many times each feature in $\Phi_a$ is seen in interaction state\replaced{. T}{, and t}his count is represented as $countF(\phi_i)$ where $\phi_i \in \Phi_i$. Then, for each $\phi_i$, we count the number of occurrences of $\eta_1$ in the game state. This count is represented as $countS(\eta_1)$. For example, in the game state shown in Fig. \ref{fig:game_interaction_a}\added{,} $countS(Wall){=}27$ and $countS(Door){=}1$. Finally, the criterion for a feature $\phi_i$ is calculated by $\frac{countF(\phi_i)}{countS(\eta_1)} \times 100$.
This division is used to normalize the $Count(\phi_i)$ and this normalization supports achieving a similar behavior in other levels. Since the algorithm is finished with this segment, it progresses the game state by applying each action in the trajectory $\tau_a$. Note that the interaction state in the game state is reset after this step. This procedure continues until every segment is processed. 

At line \ref{alg:mgp_irl:check_last_cluster}, the algorithm checks if there is a remaining segment, and if there is a segment, it is converted into a goal. Lastly, as stated in the interaction state, not every tester will interact with the same feature in the same way. Therefore, lines from \ref{alg:mgp_irl:create_feature} to \ref{alg:mgp_irl:likelihood} (not shown in the algorithm) are performed twice by changing the \textit{Method} parameter of $\phi_i$ to obtain whether the tester has a direction preference. We chose the one that yields a higher likelihood. 

\added{To sum up}\deleted{Consequently}, by splitting the trajectory using the interactions, we are able to use non-zero weights, find an estimate for the repetition of the features, calculate the direction preference of testers\added{,} and most importantly split the trajectory into policies that fall under a certain likelihood threshold. \added{Consider the game depicted in \figurename{ \ref{fig:game_interaction_a}} and consider that a human executed the following trajectory. First, she tried to go through the door, but on the way attacked the walls, and then attacked the key. MGP-IRL dissects this trajectory into interactions described as above. In the first iteration, the weights, method, repetition limit for walls are found.
% ,features are extracted and the first feature is set as the first trajectory.
In the second iteration, the algorithm considers merging the first trajectory with the trajectory that includes the door. Since the tester moved towards the goal and attacked the walls along this way, the likelihood of the combined trajectory increases. Therefore, the algorithm combines these trajectories and sets the current trajectory as the combined trajectory. In the third iteration, the algorithm considers merging it with the trajectory that includes the Key. This combination will decrease the likelihood of the trajectory as the tester can execute this action early in the trajectory. Depending on the selected $\kappa_T$, the threshold check may fail. Then, the initial combined trajectory is converted into a goal, and the remaining trajectory is converted into another goal. Finally, these two goals are inserted into the goal sequence $\mathcal{H}$ consecutively. If the threshold check passes, the whole trajectory is converted into a single goal.}

% We use linear features; therefore, we assume that for a segment of a trajectory to be sampled from an optimal policy, first its smallest chunk should be sampled from an optimal policy.
Our approach is different from \cite{Michini:2012} and \cite{Sosic:2018}, since their sub-goal definition is based on the state, \deleted{which restricts our human-like agents to play other levels}\added{which restricts the applicability of the extracted test goals to other levels}. Moreover, due to our feature definition, we should know the number of repetition and direction preference\added{s, which are not computed in their approach}.\deleted{ Lastly, rather than calling the goals in trajectories as sub-goals as in \cite{Michini:2012} and \cite{Sosic:2018}, we refer to them as goals.}
% We assume that for a trajectory to be sampled from optimal policy, first its smallest chunk should be sampled from an optimal policy.

\begin{algorithm}
  \caption{Multiple Greedy Policy Inverse Reinforcement Learning for extracting test goals from human trajectories}\label{alg:mgp_irl}
  \begin{algorithmic}[1]
  \Procedure{MGP-IRL($G$, $\tau$, $\kappa_T$)}{}
  %\BState \emph{step 1}
  \State $\zeta_{0..n}, \tau_{0..n}, n \gets \Call{SplitTrajectory}{G,\ \text{$\tau$}}$ \label{alg:mgp_irl:split}
  \State $\Phi_a \gets \{\ \}, \tau_a \gets [\ ], \kappa_a \gets 0, \mathcal{H} \gets [\ ]$ \label{alg:mgp_irl:init}
  \State $i \gets 0$
  \While {$i \leq \textit{n}$}
  %\State $a$
  \State $\phi_i \gets \Call{CreateFeature}{\zeta_i}$ \label{alg:mgp_irl:create_feature}
  \State $\tau_b \gets \Call{Concatenate}{\tau_a,\ \tau_i}$
  \State $\Phi_b \gets \Phi_a \cup \phi_i$ \label{alg:mgp_irl:combine_feature}
  \State $\Phi_b \gets \Call{AnalyzeRepetitions}{G,\ \tau_b,\ \Phi_b}$ \label{alg:mgp_irl:analyze_rep}
  \State $\Phi_b \gets \Call{IRL}{G,\ \tau_b,\ \Phi_b}$ \label{alg:mgp_irl:irl}
  \State $\kappa_b \gets \Call{CalculateLikelihood}{G,\ \tau_b,\ \Phi_b}$ \label{alg:mgp_irl:likelihood}
  \If {$\Phi_a$ is $\{\ \}$ OR $\kappa_a - \kappa_b \leq \kappa_T$ OR $\Phi_a$ is $\Phi_b$} \label{alg:mgp_irl:check}
  \State $\Phi_a \gets \Phi_b, \tau_a \gets \tau_b, \kappa_a \gets \kappa_b$ \label{alg:mgp_irl:combine_cluster}
  \State $i \gets i + 1$
  \Else
  \State $G, h \gets \Call{CreateGoal}{G,\ \Phi_a,\ \tau_a}$ \label{alg:mgp_irl:create_goal}
  \State $\mathcal{H} \gets \Call{Append}{\mathcal{H},\ h}$
  \State $\Phi_a \gets \{\ \}, \tau_a \gets [\ ], \kappa_a \gets 0$ \label{alg:mgp_irl:reset_cluster}
  \EndIf
  \EndWhile
  \If {$\Phi_a$ is not $\{\ \}$} \label{alg:mgp_irl:check_last_cluster}
  \State $G, h \gets \Call{CreateGoal}{G,\ \Phi_a,\ \tau_a}$
  \State $\mathcal{H} \gets \Call{Append}{\mathcal{H},\ h}$
  \EndIf
  \Return $\mathcal{H}$
  \EndProcedure
  \end{algorithmic}
\end{algorithm}

\subsection{Generating Test Sequences}
Testing is a rigorous agenda, and an update to the code or design require\added{s} re-testing. Thus, the testing cycle becomes more tiresome over time. Therefore, the ability to automatically create new sequences is necessary. \deleted{This section explains how new test sequences are generated using an agent. In this study, there are two types of agents, synthetic and human-like agents. Since both of these agents hold an array of goal sequences, the same algorithm can be used by both of them to generate new test sequences.}\added{ To fullfil this necessity, we propose using agents to create new test sequences automatically using test goals. This section explains how our agents generate new test sequences.}

\deleted{We use learning algorithms to generate the test sequences from agents.} Learning algorithms take an environment and a reward vector, but \replaced{our}{the described} agents contain a goal sequence. Therefore, an agent plays the goals sequentially by its feature vector, and then this sequence is checked to evaluate how much of the criteria are fulfilled. This fulfillment condition is determined by a criterion threshold $c_T$. A criterion threshold is required since the synthetic agent has no real experience and the human-like agent plays a different level.
%TODO this is repeated in synthetic agent Goal
\deleted{If the agent does not fulfill the threshold for a goal, the agent does not get to play the next goal. We chose not to progress the agent if it fails a goal for two reasons: the agent might have encountered a bug, or the learned sequence might not match the level.} Furthermore, \replaced{the criterion of each feature is}{the criteria are also} used to dampen the weights \replaced{after criterion is fulfilled}{in the feature} and to reward goal completion. This supports to distinguish various \deleted{agents}\added{test goals} that have distinctive criteria but similar features. Moreover, the main objective of the agent is to complete its \added{test }goal\replaced{. As}{, as} goal completion depends on criteria, \added{we reward the agent w.r.t. the }completion percentage \added{of the goal}\deleted{is employed to reward the agent}. In this study, this additional reward is \replaced{defined}{calculated as} $reward^{completion}$ where \added{$reward$ is a positive real number}\deleted{$\{reward \in \mathbb{Q}~|~reward \geq 0\}$} and $\deleted{\{completion \in \mathbb{Q}~|~}completion \in [0, 1]\deleted{\}}$. Individual completion is calculated for each criterion and then these values are multiplied to obtain the total completion\added{ of the current goal.}\deleted{, t}\added{ T}herefore, if a single criterion's completion \added{is }calculated as zero, then the total is zero as well. Note that after each goal is completed, the current interaction state is reset.

\added{
We use the state space, action space, and reward function described in Section \ref{sec:methodology:is}.  We have two kinds of agents: MCTS and RL agents. In our MCTS agent, we use knowledge based evaluations in the MCTS to evaluate the states in simulation phase. We use transpositions \cite{Childs:2008} to share information amongst states, and utilize UCT3 \cite{Childs:2008} in selection phase of MCTS. In our RL agent, we use Sarsa($\lambda$) algorithm described in Section \ref{sec:pre:rl}. Eligibility traces support propagating the calculated long term reward, which is obtained by fulfilling a goal. Lastly, we chose Boltzmann exploration policy \cite{Kormelink:2018}.
}

\deleted{As the main learning algorithm, Sarsa with Boltzmann exploration policy \cite{Kormelink:2018} is chosen. We integrated eligibility traces TD($\lambda$) \cite{Singh:1996}, Sarsa($\lambda$), which helps propagating the calculated long term reward. This approach is referred to as an agent using RL. Lastly, for MCTS, we use UCB1, implemented transpositions and move groupings \cite{Childs:2008} to MCTS as we intend to shrink the state space extended by interaction state.}

% \paragraph{Test Oracle}
\textbf{Test Oracle:} Generating and executing a test sequence are not enough without determining whether the test fails or not. Test oracle provides a mechanism that determines whether the software behaves as expected in a test run. Automating the test oracle greatly improves the test execution due to the elimination of manual examination of a test execution result. As our aim is not detecting visual glitches but to find the dissimilarities between the game design and the implementation, we did not use a vision-based oracle.\deleted{ In addition, a vision-based oracle \cite{Lovreto:2018} would be too costly, in terms of time, to check the state of the game grid during playing the test sequences.} \replaced{Therefore, we opt to use a model-based oracle that determines fail and success of an execution at each iteration of the game loop}{Therefore, we opt to use a model-based approach that exploits the game loop for bug checking} \cite{Varvaressos:2017}.

\added{This oracle makes a model comparison using the game scenario graph and the game. This comparison verifies the game transitions using the scenario graph transitions. For example, if the avatar does not possess the key and the door is in play, the game should not be won. However, the model from the game scenario graph cannot be used to catch bugs such as wall collision. Thus, we used additional constraints to catch these bugs such as if the avatar dies, it should have collided with the fire sprite; or the position of avatar should not overlap with any wall. The oracle checks if the game state and interactions violate any of the constraints. The model and the constraints are given by the game developer.}
\deleted{
This oracle has global rules such as a movable sprite should not occupy the same position with an immovable sprite (unless it is the \textit{Floor}). There are also game-specific rules such as if the avatar does not possess the key and the door is in play, the game should not be won; if the avatar died, did it interact with harmful sprites. These rules, which are checked at the end of the game loop, enables the oracle to verify the game state using global rules and interactions with using specific rules.}

\section{Experiments}\label{sec:experiments}

\subsection{Experimentation Setup}
In this study, three different games are prepared using GVG-AI framework \added{(see \appendixname{ \ref{sec:appendix_2} and \ref{sec:appendix_3}})}. These games have varying difficulties and their dimensions are 6x7, 8x9, and 10x11. \added{We refer to these games as Game A (6x7), Game B (8x9), and Game C (10x11), respectively. }Each of these games has four different levels. These levels \replaced{differ}{are altered} in \replaced{terms of layout}{sprite positions but have the same sprite set}. \added{Also, the avatar can use the \textit{Use} action only in the first two levels. }6x7 game is quite simple\replaced{:}{,} the avatar has to pick up the key and go to the door. In the 8x9 game, the avatar has to extinguish the fire, pick up the key, and finish the level by going through the door. In the 10x11 game, the avatar has to create the key by combining the key parts, pick up the key, and finish the level by going through the door. In the 8x9 game, each level has a different game scenario graph.\deleted{ We refer to these games as Game A (6x7), Game B (8x9), and Game C (10x11), respectively.}

\added{
We use fault seeding to verify our approach. Fault seeding is a technique to evaluate the fault detection rate of software tests and test process \cite{Ammann:2008}. The source code of our games are the VGDL descriptions ---not the GVG-AI engine; hence we altered VGDL descriptions while inserting faults. These faults affect the implementation of the game to behave different than the ideal design of the game. We use \cite{Lewis:2010} as a reference to diversify the bugs. We alter the VGDL game description by: removing lines from interaction set, changing the order or the name of the sprites in the interaction, and adding fallacious interactions.
We inserted a total of 45 bugs into the VGDL scripts of the games. During scoring, if there are multiple occurrences of the same bug, it is counted as one.
% Fault seeding is … 
% Programcinin olasi yapacagi hatalar. Source code bizde VGDL.
% Taxonomy ref.
% Bizde engine test edilmiyor

% We chose to modify the VGDL for inserting bugs as it is elegant for this task. The simplest examples are deleting a line from the VGDL, swapping sprite names, reversing termination conditions. Furthermore, there are bugs such as adding \textit{FakeWalls} which are destroyed after an action. For these bugs, we grouped these sprites, for example \textit{NormalWalls} with \textit{FakeWalls} under the name \textit{Wall}. Since human testers cannot distinguish different walls, the agents should also. Moreover, we added bugs such that the result of an interaction depends on the direction of the interaction. For these bugs, we created new interaction effects. These enable us to insert bugs such as not being able to pick up the key from certain directions or ability to pass from flames. Lastly, these bugs can be further diversified by including the state of the avatar.
}

We collected a total of 427 trajectories from 15 different human participants that have different gaming and testing experience. During the data collection, for each game, we stated the rules of the game and their goals, but there were no directions. Testers were able to play the same level any number of times and could even go back and forth between levels and games. There were tutorial levels for players to get used to the controllers and the environment. \added{We should state that except the tutorial levels, all of the games included bugs.} We used the GVG-AI framework to collect trajectories, and for each game, a total of 118, 173, and 136 trajectories are collected, respectively. It should be noted that 8x9 has \replaced{more game scenario paths than the other games}{the most complex game graph}; as a result, testers executed more tests on this game.

We applied MGP-IRL to these trajectories and used Maximum Likelihood Inverse Reinforcement Learning (MLIRL) \cite{Babes-Vroman:2011} as the IRL algorithm since it internally calculates the likelihood of the trajectory and it is robust to slight mistakes or noise. We chose three different likelihood thresholds $\kappa_T$: 0.0, 0.5 and 1.0, note that MLIRL calculates log likelihood. \added{We compared these three different threshold values to assess the effectiveness of the MGP-IRL. Threshold of 0.0 is our proposed approach for finding weights and threshold of 1.0 is close to using MLIRL.} We used the following parameters for MLIRL described in \cite[Algorithm 1]{Babes-Vroman:2011}: $M{=}20,~\beta{=}5,~T{=}0.01$. For each game, for each tester, and for each level, \deleted{an agent is trained}\added{a human-like test goal is learned} using the collected trajectories from the other three levels. Thus, \deleted{this technique}\added{human-like agents} generate\deleted{s} more tests than the original human tester.

For the synthetic \deleted{agent}\added{test goals}, we entered the game scenario graph and the sprites. \deleted{The synthetic agent generated main scenario paths and added modifications to them.} For \replaced{our 6x7, 8x9 and 10x11}{these} games, synthetic agent produced 28, 234, and 88 different test sequences, respectively. We used all path coverage since these game scenario graphs do not contain any loops. Lastly, to investigate the effect of modifications on bug finding, a baseline agent is created. \deleted{This agent is similar to the synthetic agent but does not use the modifications.} \added{The baseline agent uses test goals directly generated from the graph. 

We did not use off the shelf testing tools record/replay, test automation frameworks, and monkey testing in our comparison. The record/replay tools will fail even when the direction of the avatar is different. Test automation frameworks require an expert to manually design scenarios, which is not only arduous, but also not scalable. Lastly, monkey testing generates random events to stress test the UI, which is not our test objective. Therefore, COTS testing tools were not adequate in our experiments.}

We chose the following parameters for \deleted{our RL agent}\added{Sarsa($\lambda$)}: $\gamma{=}0.95,~\beta{=}1$, temporal difference $\lambda{=}0.8$, learning rate $\alpha{=}0.03$, and for\deleted{ our} MCTS\deleted{ agent}: $\gamma{=}0.95$, exploration term $C_p{=}3$, \deleted{8 rollouts}\added{rollout depth is 8}, 300ms for computation budget on i7-4700HQ using single core. \deleted{Moreover, w}\added{W}e ran the MCTS \deleted{agent 20}\added{5} times\deleted{ to obtain a justifying result}. Our criteria threshold $c_T$ was set to 0.01, goal completion reward was set \replaced{to}{as} 10, and the features that have other than non-zero reward are considered as a singular feature with a reward of -1. \added{We have chosen not to progress an agent if the agent does not accomplish its current goal. }\deleted{Nevertheless, a}\added{A}s there is no clear definition of a terminal state, we experimented with game lengths 50, 100, 150, 200, 250, and 300 and chose the one that achieved the highest criterion completion. For Game C (10x11), we set direction preference to \textit{All} only to decrease the memory requirements. \added{The running time of Sarsa agent was between a few minutes to 6 hours, depending on the complexity of the goal sequence being played. }Lastly, some bugs allow the testers to go out of the intended grid area\added{. S}\deleted{ s}ince outside of the grid was not modeled, \deleted{and} we assumed that the agent was interacting with the \textit{Floor} sprite. While training with MCTS and RL, this behavior caused problems such as divergence. Therefore, we assigned a negative reward when the agent tried to explore outside after getting out of the map.

\deleted{We inserted 40 bugs to these games. Bug injection is done mostly by changing the VGDL code. We used \cite{Lewis:2010} as a reference to diversify the bugs. During scoring, if there are multiple occurrences of the same bug, it is counted as one.}

\subsection{Results}

In this study, we asked the following questions.
\begin{itemize}
  \item Which \deleted{agent}\added{test goal} technique is better in finding bugs?
  \item \added{What is the difference of MCTS and Sarsa agents in bug finding?}
  \item \replaced{Which human-like agent is more similar to human testers?}{Can the human testing behavior be captured and can this behavior be re-targeted to similar levels?}
\end{itemize}

To assess bug finding performance, we compared \replaced{eight}{four} different tester groups: original human testers, \added{three human-like Sarsa agents, one MCTS agent with likelihood threshold of 0.0, one baseline Sarsa agent, and one MCTS and one Sarsa agent with synthetic goals.}\deleted{three human-like agents, one baseline agent, and one synthetic agent.} There are two different figures for bug finding performance\replaced{. T}{, t}he first one uses barplot to compare these groups in each game. Individual agents contribute to the score\added{ of a group} together,\deleted{ hence in barplot,} \added{and }the total bug count is considered as total unique bugs.
\added{The bugs found percentage of MCTS agents is mean of 5 runs.}
The second\added{ figure} compares the individual testers in human-like agents and the original human testers. 
\added{We use violin plot to depict this figure.}
\deleted{This figure uses the violin plot to depict the distribution and dots for the individual testers.}

% In each experiment we compare the individual performance of testers in human-like agents versus individual human testers, contrast the human-likeness of each human-like agent, show the action length of human, human-like agent with likelihood threshold of 0.0, synthetic and baseline, and show number of splits done by MGP-IRL.

Human-like agents' similarity is evaluated using the cross-entropy of human behavior and agent behavior. The trajectory obtained from the human tester is replayed to find a list of interactions $\zeta^h_{0,...,n}$, then the trajectory executed by a human-like agent that learned from this human tester's trajectory is replayed to find a second list of interactions $\zeta^a_{0,...,n}$.
Using \deleted{the }$\eta_0, \eta_1, Type, Avatar_{State}$, each list is binned, and the frequency of each bin is used to obtain the cross-entropy of $\zeta^h_{0,...,n}$ and $\zeta^a_{0,...,n}$. We removed the position and direction components of interactions \replaced{during comparison}{while comparing} as these are highly dependent on the level layout. Lastly, although these do not hold comparison values, the number of actions performed by the agents and the number of splits performed by MGP-IRL are examined. Action length figures are shown using violin plots. Cross-entropy and number of splits figures are shown using box plots with IQR=1.5, except cross-entropy uses log scale \cite{Matplotlib}\cite{Seaborn}.

\begin{figure}[]
  \centering
  \includegraphics[width=1.0\columnwidth]{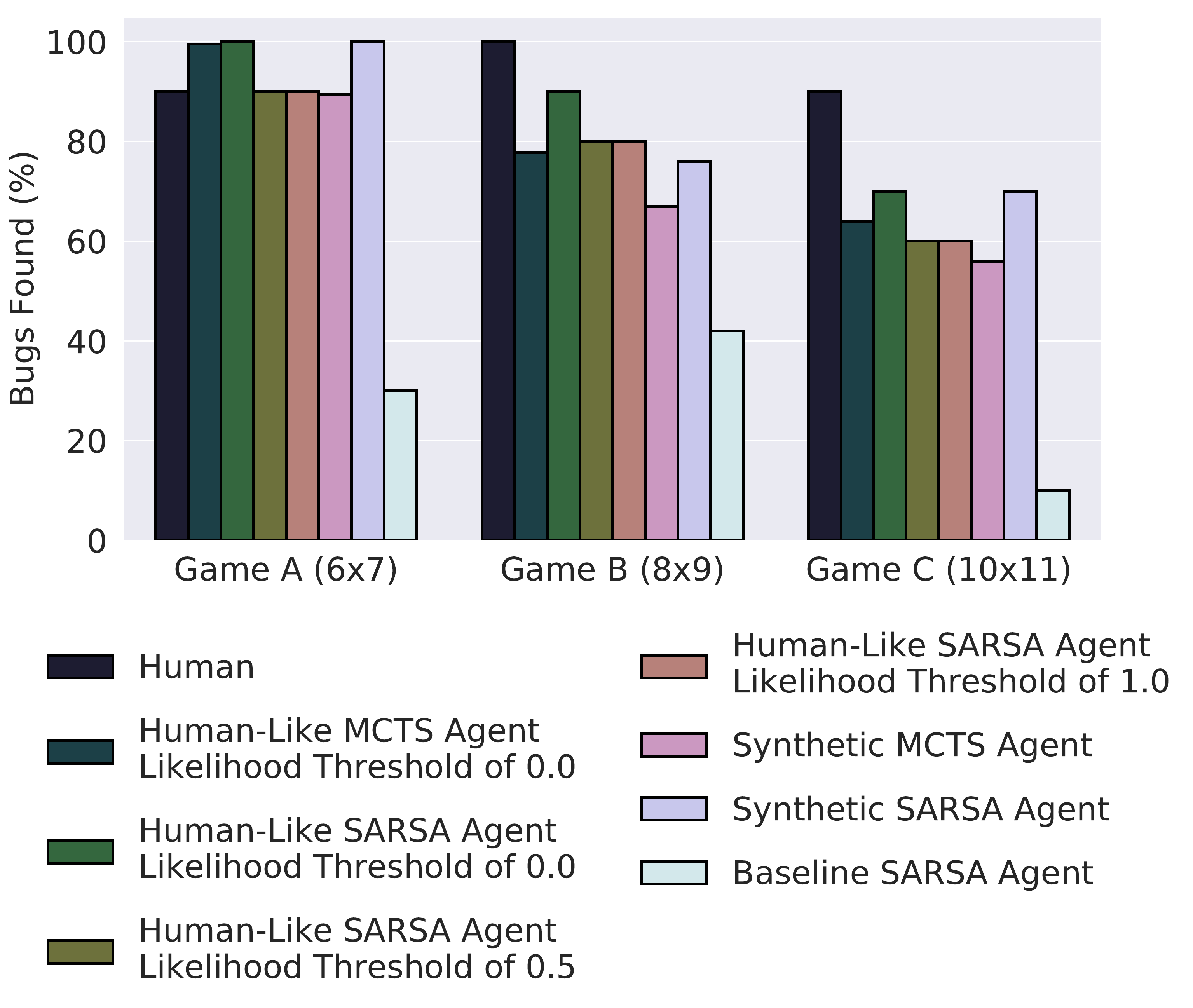}
  \caption{The Percentage of Bugs Detected by Human Testers and The Generated Agents For Each Game Under Test}
  \label{fig:total_bugs_found}
\end{figure}

\begin{figure*}[]
  \centering
  \includegraphics[width=1.0\textwidth]{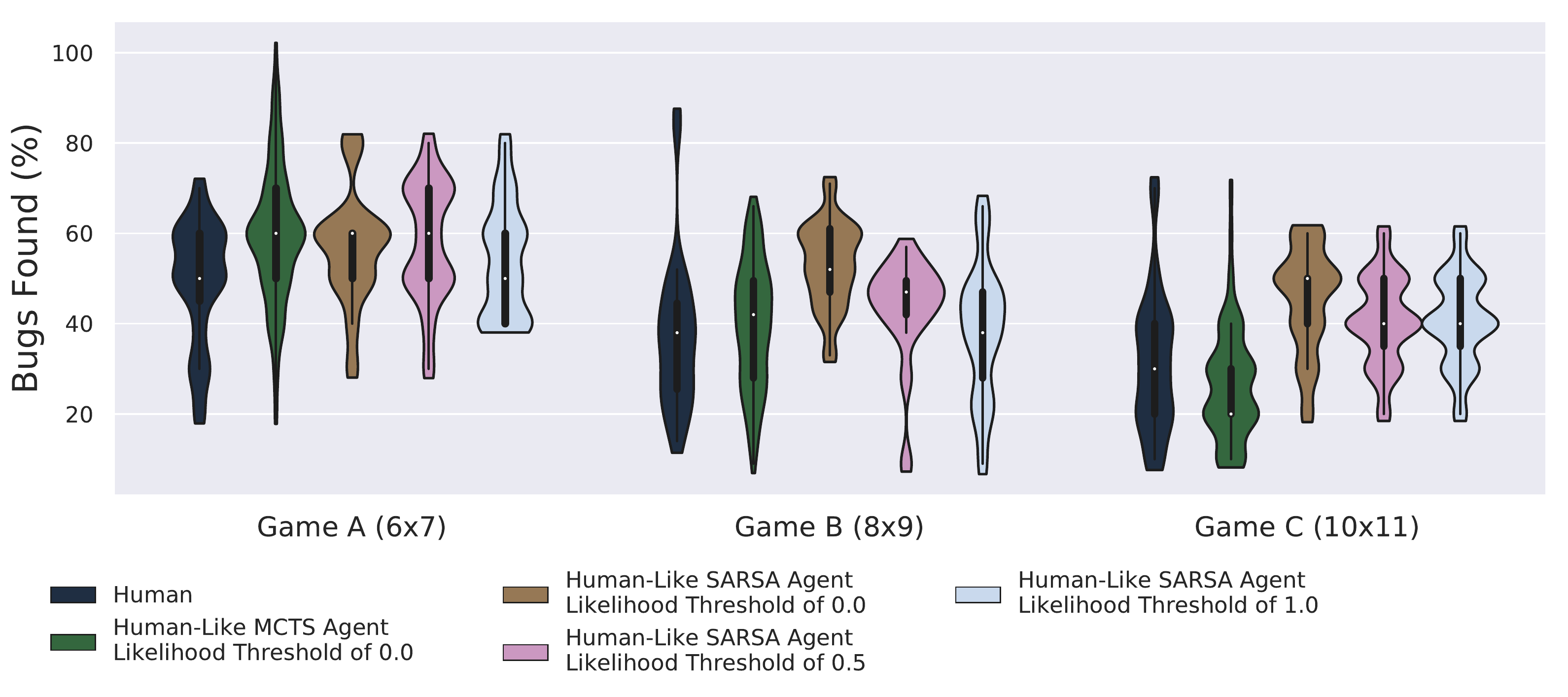}
  \caption{Percentage of Bugs Detected by Human Testers and The Human-Like Agents For Each Game Under Test}
  \label{fig:individual_bugs_found}
\end{figure*}

\begin{figure*}[]
  \centering
  \includegraphics[width=1.0\textwidth]{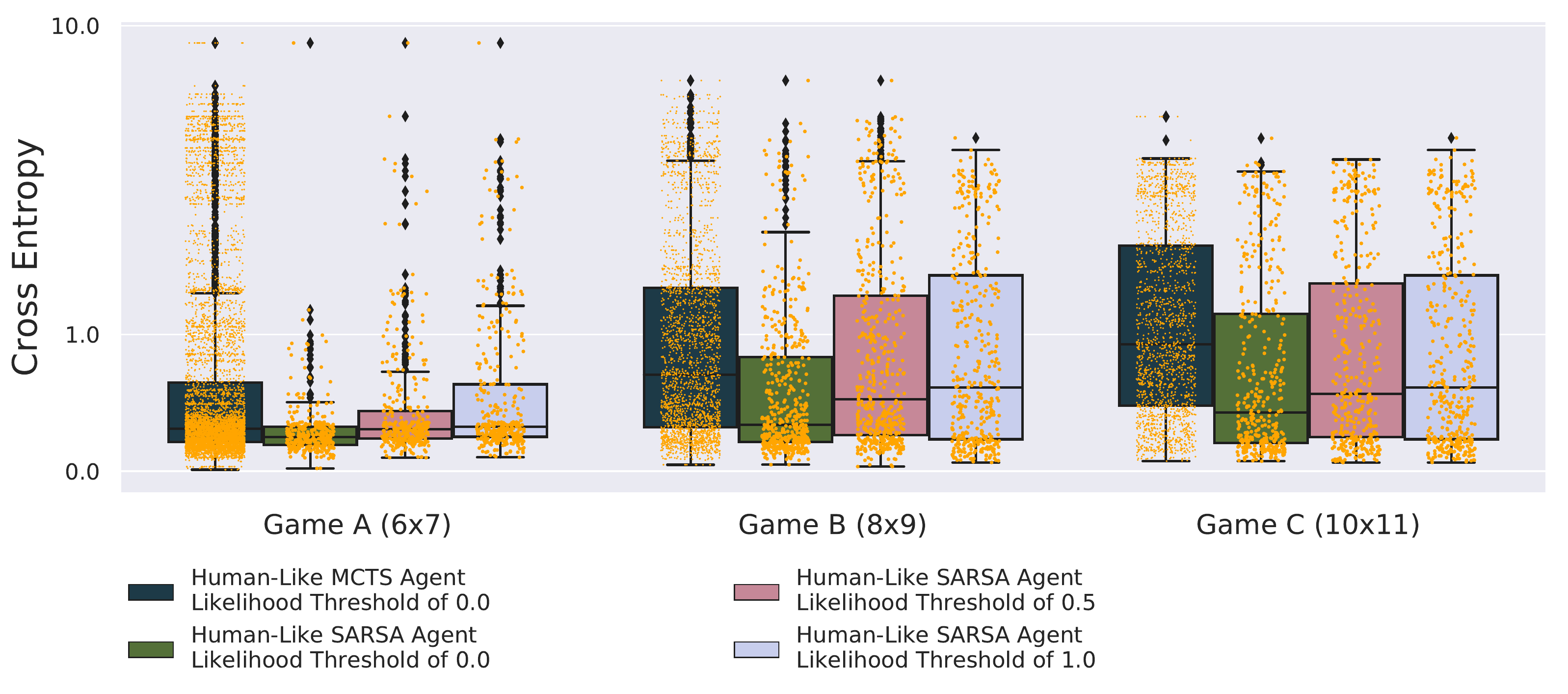}
  \caption{Cross Entropy of Human-Like Agents For Each Game Under Test}
  \label{fig:cross_entropy}
\end{figure*}

\begin{figure*}[]
  \centering
  \includegraphics[width=1.0\textwidth]{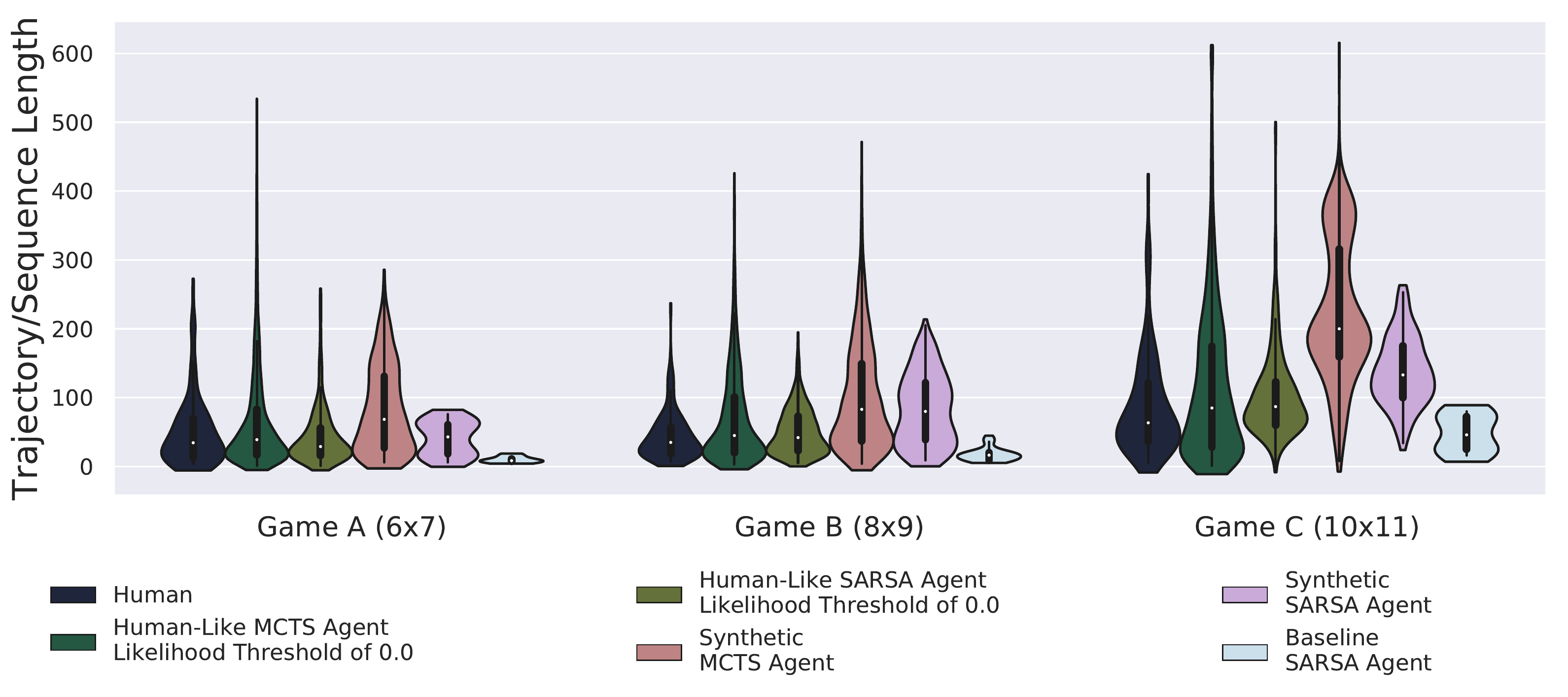}
  \caption{Trajectory/Sequence Length of Human Testers and The Generated Agents For Each Game Under Test}
  \label{fig:sequence_length}
\end{figure*}

\begin{figure*}[]
  \centering
  \includegraphics[width=1.0\textwidth]{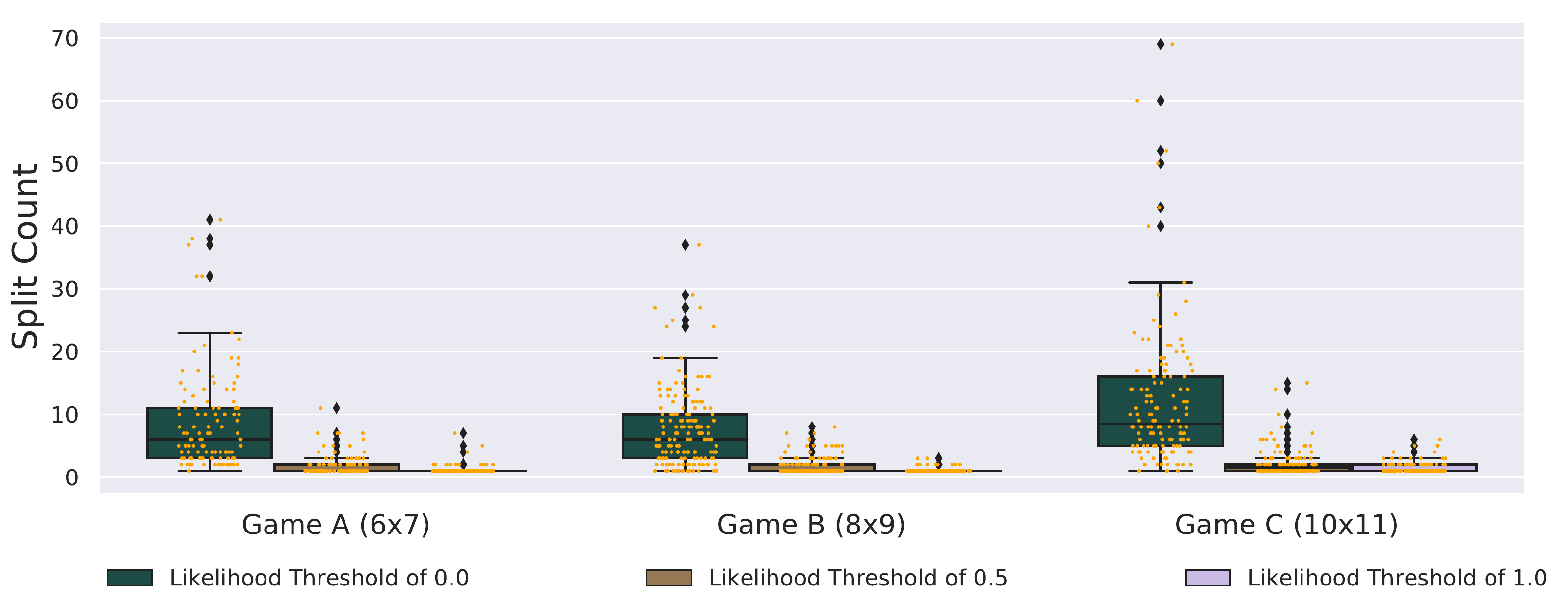}
  \caption{Number of Splits Done by MGP-IRL of Trajectories Collected From Game A (6x7), Game B(8x9), and Game C(10x11)}
  \label{fig:number_of_splits}
\end{figure*}

Fig. \ref{fig:total_bugs_found} shows the bug finding performances of various agents in each game. \replaced{We see}{It can be seen} that humans found all of the bugs in \deleted{two of the games}\added{the second game} and found 90\% of the bugs in \deleted{Game A (6x7)}\added{the first and the third game}. In \deleted{this game}\added{Game A (6x7)}, the human-like agent\added{s} with likelihood threshold of 0.0 and the synthetic \added{Sarsa} agent surpassed the human performance and found all the bugs\added{, and the synthetic MCTS agent \replaced{showed}{obtained} a similar performance to of humans}. Moreover, there is a clear difference between the baseline agent and the synthetic agent. The bugs found by baseline agent can be interpreted as number of bugs found by playing only the scenarios specified by the designer. The overall performance difference between the MCTS agent and \replaced{Sarsa varies from}{RL is} 5\% to 10\%\deleted{, however, in different runs, \replaced{we observe}{it can be seen} that MCTS can reach the performance of RL}. In Game B (8x9) the difference between different testers is evident. Human-like agents surpassed the synthetic agent and there is a 10\% difference between the best human-like agent and the human testers. \deleted{Nonetheless, the performance of the baseline increased significantly.}\added{The performance of the baseline agent is over 40\%}. In Game C (10x11), we can see the gap between the humans and agents increase, \added{and }the synthetic \added{Sarsa} agent's performance is on par with the human-like \added{Sarsa} agent with likelihood threshold of 0.0. \added{Lastly, test goals used by MCTS agents find fewer bugs in all three games.}
% Lastly, for MCTS agents we observed the following standard deviations, human-like agent with likelihood threshold of 0.0: 2.67, human-like agent with likelihood threshold of 0.5: 4.87, human-like agent with likelihood threshold of 1.0: 8.56, synthetic agent: 7.30, baseline agent: 5.00.

% ‘. Experience has shown that recorded tests can’t survive the system change. Logs, produced by recording software, are unmanageable and require to be re-recorded on every system update, thus making automated testing literally useless. Our decision was to write test code by hand. All we needed now is to choose the appropriate programming language suitable for the task.’
% >>> MCTS for all games? Petri nets similar to our baseline agent?
% >>> Record and Replay
% >>> Monkey Testing
% >>> Human Labor azaltmak staff month labor decrease

% \subsubsection
{\deleted{1) Experiment 1: Agents using MCTS Testing Game A (6x7)}}
\deleted{
In Fig. \ref{fig:individual_bugs_found} (Game A (6x7) MCTS) we see that two of the human-like agent testers with likelihood threshold of 0.0  found all of the bugs, exceeding the RL agent, and the mean performance is almost the same as that of RL. In Fig. \ref{fig:cross_entropy} (Game A (6x7) MCTS) the noise induced from the stochasticity of MCTS can be seen compared to the game played with RL. In Fig. \ref{fig:sequence_length} (Game A (6x7) MCTS) it can be seen that all MCTS agents executed more actions than the RL agents.
}

\subsubsection{Experiment \deleted{2}\added{1}: Agents \deleted{using RL} Testing Game A (6x7)}

Fig. \ref{fig:individual_bugs_found} (Game A (6x7)\deleted{ RL}) depicts the scores of different human-like agents in finding bugs. In a simple game like this, where an agent can generate almost perfect runs, \replaced{we see}{it can be seen} that \replaced{the}{their} individual performance \added{of human-like agents }is at least the same \added{as }or better than individual humans. In Fig. \ref{fig:cross_entropy} (Game A (6x7)\deleted{ RL}), it is evident that human-like \added{Sarsa} agent with likelihood threshold of 0.0 performed the most similar interactions\deleted{ amongst three}. Moreover, the likelihood threshold affected this similarity in a descending pattern.\deleted{ Nevertheless, there is a sequence that neither these three methods were able to learn.} In Fig. \ref{fig:sequence_length} (Game A (6x7)\deleted{ RL}), \deleted{it can be seen that most of the testers are shaped like teardrops, but the synthetic agent. T}\added{t}he human-like \added{MCTS} agent with likelihood threshold of 0.0 performed most actions among all agents. Baseline agent was able to finish the game in less than 15 actions. Fig. \ref{fig:number_of_splits} (Game A (6x7)\deleted{ RL}) displays the number of splits resulted from MGP-IRL\added{: }\deleted{, human-like agent with} likelihood threshold of 0.0 split more than the others, \deleted{human-like agent with} likelihood threshold of 1.0 agent split the lowest and most of the times, it considered the trajectory as a whole. \added{Lastly, the total bugs found ---of all 5 runs--- of human-like MCTS and synthetic MCTS are 100\%.}

\subsubsection{Experiment \deleted{3}\added{2}: Agents \deleted{using RL} Testing Game B (8x9)}

In Fig. \ref{fig:individual_bugs_found} (Game B (8x9)\deleted{ RL}) human-like agents \replaced{have a higher bug finding mean than}{increased the overall performance} of individual human testers, but they were not able to perform on a par with the best human tester. Cross-entropy of interactions are similar \added{amongst agents }in this game with the human-like agent with likelihood threshold of 0.0 leading, which can be seen in Fig. \ref{fig:cross_entropy} (Game B (8x9)\deleted{ RL}). Since one of the sprites was missing in the last level of this game, retargeting was not ideal, and the overall cross-entropy is higher compared to Game A (6x7). In Fig. \ref{fig:sequence_length} (Game B (8x9)\deleted{ RL})\added{,} we see that the synthetic \added{MCTS} agent executed most actions among all agents and had a distinct shape than all other agents. The baseline agent executed at most 40 actions. Fig. \ref{fig:number_of_splits} shows that (Game B (8x9)\deleted{ RL}) human-like agent with likelihood threshold of 1.0 performed the least amount of splits while human-like agent with likelihood threshold of 0.0 divided the trajectory the most. \added{Lastly, the total bugs found ---of all 5 runs--- of human-like MCTS is 90\% and synthetic MCTS is 76\%.}

\subsubsection{Experiment \deleted{4}\added{3}: Agents \deleted{using RL} Testing Game C (10x11)}

As seen in Fig. \ref{fig:individual_bugs_found} (Game C (10x11)\deleted{ RL}), \deleted{likelihood}\added{human-like Sarsa} agents improved the overall performance of individual agents, except one tester. \added{Human-like MCTS agents has lower mean values compared to all testers.} Fig. \ref{fig:cross_entropy} (Game C (10x11)\deleted{ RL}) shows that the mean of cross-entropies \added{Sarsa agents} are below 0.5 and quite alike when compared to other games. In Fig. \ref{fig:sequence_length} (Game C (10x11)\deleted{ RL}) we see that all of the agents executed more actions than previous games. Fig. \ref{fig:number_of_splits} (Game C (10x11)\deleted{ RL}) reveals that the number of splits for each agent increased when compared to other games. \added{Lastly, the total bugs found ---of all 5 runs--- of human-like MCTS 90\% and synthetic MCTS are 80\%.}

\section{Discussion}\label{sec:discussion}

In this paper, we presented a technique for \replaced{capturing}{understanding} tester behavior, namely interaction state, and introduced two different strategies to generate \deleted{a tester agent}\added{test goals for agents}: synthetic and human-like. We compared the bug finding performance of the\deleted{se} agents in three different games and evaluated the similarity of the human-like agents with the original human testers.

Interaction state helped to distinguish previously equivalent states. Interaction state supports this behavior and many more such as attacking all of the walls, covering all empty spaces. Consequently, we were able to model the testing behavior using MDP and MGP-IRL was able to learn tester heuristics from collected trajectories. Note that, these trajectories were collected from games that contained bugs. Though we used the interaction state primarily for testing, it can benefit the gameplay. There can be many hidden doors and other rare objectives in a game, and an agent utilizing the interaction state can engage with these objectives.

Creating a synthetic \deleted{agent}\added{test goal} out of game scenario graph and inserting modifications were valuable since the synthetic agent beat baseline in every experiment (Fig. \ref{fig:total_bugs_found}). Baseline agent surpassed half of the \added{individual} human testers in Game B (8x9) (Fig. \ref{fig:total_bugs_found} and \ref{fig:individual_bugs_found}), which has the most complex game scenario graph. This \added{fact} reveals that these testers were not able to cover all intended paths of the game, as it is difficult to understand the underlying graph information.\deleted{ Our baseline agent is comparable in behavior to Petri nets \cite{Becares:2016}, but sequential approach supported modifications.} Graph coverage provided the path to play the game and modifications guided the agent to numerous paths. Game graph has several advantages as it makes the agent to play each intended scenario and guarantees that these paths are covered, unlike in \cite{Pfau:2017}. Moreover, the generated modifications encouraged the agent to stress the limits of the game. In the experiments, synthetic agent beat every individual tester in human-like agents and most of the human testers (Fig. \ref{fig:total_bugs_found} and \ref{fig:individual_bugs_found}), which demonstrates the potency of the synthetic agent. Lastly, it provides a flexible mechanism with modifiable coverage criterion and number of modifications to conduct tests without collecting huge amounts of data. \added{However, synthetic test goals are better utilized by the Sarsa agent compared to MCTS agent, as it found more bugs in all three games (Fig. \ref{fig:total_bugs_found}).
}

% \added{ Implementing this tester distinction in synthetic agent can be achieved from applying different tester personalities described in \cite{Romero:2018}.}

Human testers individually were not able to find all of the bugs nor surpass the synthetic agent, but when their performance outcomes were combined, they were able to find most of the bugs and exceed the synthetic agent in \replaced{two}{three} out of \replaced{three}{four} experiments (Fig. \ref{fig:total_bugs_found}). This situation is also seen in human-like testers\replaced{ in which}{,} the best human-like agent performed the same or better than the synthetic agent in every experiment (Fig. \ref{fig:total_bugs_found}). We observe that different human testers traversed different paths of the game and revealed different bugs. Therefore, it is crucial to find distinct testers. Moreover, as human-like \replaced{test goals were extracted from}{agents were trained using} these testers' sequences, they benefited from this variance as well. When we compare the bug finding performance of different human-like agents, \replaced{we observe}{it can be seen} that human-like \added{Sarsa} agent with likelihood threshold of 0.0 is leading both in individual performance (Fig. \ref{fig:individual_bugs_found}) and in overall performance (Fig. \ref{fig:total_bugs_found}). We can attribute this success to multiple goals approach. There are three reasons for this: first, a simple goal is easier to play than a more complex one; second, verifying one goal at a time is better when an agent plays another level since level composition may cause the agent to prematurely skip a feature. Third, the order of test steps in a test sequence is important as this order is designed for a purpose.\deleted{ Splitting goals supports preserving the order in a test sequence.}

We proposed MGP-IRL to create human-like \replaced{test goals}{agents} given tester trajectories. \deleted{MGP-IRL separates the trajectories depending on the interactions done and combines these partitions depending on the likelihood estimate. The algorithm generates a goal sequence which is based on features and criteria. }\deleted{Our goal completion definition is different from \cite{Michini:2012} and \cite{Sosic:2018} since it is based on features and criteria and it can be applied to different levels. }\deleted{Moreover, using MLIRL \cite{Babes-Vroman:2011} internally, it supports learning from sequences that are near-optimal. }In \replaced{all}{four} of the experiments we see that (Fig. \ref{fig:cross_entropy}) human-like \added{Sarsa} agents with likelihood threshold of 0.0 were able to execute \replaced{more similar interactions}{the interactions that are most similar} to human testers. We noticed that, as the likelihood threshold increases, the mean cross entropy also increases. Bug finding performance of these agents has a non-increasing pattern with the increase of likelihood threshold (Fig. \ref{fig:individual_bugs_found}). Therefore, we can state that human-like agent with likelihood threshold of 0.0 is both the most human-like and the most successful agent in finding bugs. Fig. \ref{fig:number_of_splits} shows the number of times the MGP-IRL split the trajectory. In all of the games, likelihood threshold of 1.0 extracted the weights of the features using the whole trajectories.\deleted{ Thus, we can state that, this approach is similar to applying MLIRL \cite{Babes-Vroman:2011} and its bug finding score is decent (Fig. \ref{fig:total_bugs_found}).}

We used \replaced{Sarsa}{RL} and MCTS\added{ agents} to generate test sequences. Rewards obtained from certain interactions \replaced{led}{lead} the agents to accomplish test goals. The accomplishment of test goals is evaluated using criteria and if the agent successfully fulfilled these criteria\added{,} then the agent played the next test goal in the sequence. This approach guided our agent to examine multiple test goals. The mean bug finding performance of \replaced{Sarsa}{RL} is greater than that of MCTS (Fig. \ref{fig:total_bugs_found})\deleted{, but agents using MCTS achieved the same performance as agents using RL except for the synthetic agent}. \deleted{The synthetic agent was not able to find all of the bugs in any of the 20 runs (Fig \ref{fig:total_bugs_found}). }\added{Synthetic MCTS agent found least amount of bugs amongst all agents.} Our manually arranged weights were more fit for \replaced{Sarsa}{RL} agent rather than MCTS. On the other hand, one of the human-like \added{MCTS} agents with likelihood threshold of 0.0 was able to find all of the bugs, which was not the case for this agent when using \replaced{Sarsa}{RL} (Fig. \ref{fig:individual_bugs_found} (Game A (6x7)\deleted{ MCTS}). After careful examination of the bugs, we noticed that the reason behind this difference was due to some fake walls. Since the human agent did not investigate all of the walls, due to stochastic nature of MCTS, this agent was able to detect this bug in some runs. \added{Moreover, when we accumulated the unique bugs found in all MCTS runs, human-like MCTS agents can find 90\% of the bugs in the third game, which is the same as humans'.} Therefore, the stochasticity of MCTS \replaced{is beneficial}{has a benefit} in testing.

\replaced{Sarsa}{RL} and MCTS have inherent bug discovering mechanisms if the agent is guided with proper goals and with right features. In the first experiment (Fig. \ref{fig:total_bugs_found} Game A (6x7)\deleted{ MCTS and RL}), due to the exploration factor of these algorithms, baseline agent outperformed one of the human testers. If the agent's goal is picking up the key and it is only possible, due to a bug, after attacking the door, then the agent can find this exact sequence. Trajectory plots reveal a difference between testing and game playing ---which is performed by the baseline agent. Game A (6x7) has a small board, but there are human testers that executed more than 100 actions (see Fig. \ref{fig:sequence_length} Game A (6x7)\deleted{ RL}). In the same game, baseline agent performs at most 15 actions, which \replaced{is}{represents} the path length\deleted{s} of game scenario graph. \deleted{When this game is played using MCTS (see Fig. \ref{fig:sequence_length} Game A (6x7)\deleted{ MCTS}), the number of actions increases in all agents.}\added{In all experiments, MCTS agents executed more actions than Sarsa agents (see Fig. \ref{fig:sequence_length}).} This is expected as \replaced{Sarsa}{RL} optimizes the whole sequence while MCTS had to pick an action in 300ms. \added{Hence, the cross entropy results of MCTS are lower in all three experiments (see \figurename{ \ref{fig:cross_entropy}}).} The difference between testing and game playing exists in other games as well (see Fig. \ref{fig:sequence_length}). Another striking factor is that the shape of the synthetic agent is quite different from human testers', but the shape of human-like testers resembles human testers. This reveals that our synthetic approach was \deleted{indeed} non-human\added{, as expected}.

\textbf{Limitations \& Challenges:} The interaction state increased the \added{number of states explored by our agents, which creates an issue for tabular RL}\deleted{memory requirement of training an agent using tabular RL}, though interaction state is considerably simple\added{ and requires less than 2KB of space for Game C (10x11)}. This \deleted{memory} problem can be solved using a function approximator. MCTS can face the same problem only if it utilizes the previously generated game tree.

The main factor affecting human-like performance is the MGP-IRL algorithm. Its greedy approach can be improved with dynamic programming, but this approach will increase the amount of time required to create \replaced{a test goal}{an agent}. On the other hand, we can direct the human tester to test a game and let the tester segment the trajectory, and then let MGP-IRL find repetition count, direction preference and rewards of the features. However, this approach will apply to an in-house testing rather than open beta-testing. Also, it should be noted that weights found by IRL\deleted{ \cite{Abbeel:2004}\cite{Babes-Vroman:2011}} get better with the amount of trajectory it processes. This \added{approach }requires the tester to repeat the same objective in different runs. Moreover, when a tester finds a bug, she will exploit it. When MGP-IRL tries to structure these trajectories, it generalizes these exploited sequences to all situations. This \added{generalization }causes a failure in learning the tester behavior. For example, if a tester finds a wall that allows the avatar to pass, the tester tries to carry other objects through there, but human-like agents interpret this wall as any wall. Consequently, the agent will test \replaced{any}{an indifferent} wall for this interaction. Moreover, Game C (10x11) consists of a free puzzle unlike Game B (8x9); hence, testers had various solutions to this puzzle. This behavior was neither captured nor repeated easily. This is a limitation of using linear features.

\deleted{Lastly, we were uncertain about whether we could capture tester behavior. Therefore, we started with simple levels and then included puzzles. However, vanilla MCTS struggled with puzzles in Game B (8x9) and C (10x11). Therefore, we present MCTS results only for Game A (6x7) which does not include any puzzles.}

\deleted{On the other hand, RL did not struggle with puzzles. In addition, w}\added{W}e chose movable sprites over enemies, because tester agents would check whether an enemy's interactions are correct with different sprites and would probably restart the test until they observe the desired behavior. Nevertheless, this behavior is not relatable to a human tester.
% Second, the bugs introduced by enemies such as artificial stupidity, cannot be understood with simplistic oracles.

\section{Conclusion}\label{sec:conclusion}

This paper focused on the problem of creating tester agents. In this regard, we proposed interaction state to capture\deleted{ tester behavior} and execute tester behavior. Furthermore, we presented \replaced{two approaches to generate test goals:}{two tester agents:} synthetic and human-like\deleted{ agents}. The synthetic \deleted{agent tests the game using the} \added{test }goals\added{ are} based on sequences from the game scenario graph. These goals are further modified to examine the effect\added{s} of unintended game transitions. Human-like \replaced{test goals are}{agents} learned from human testers' collected trajectories using MGP-IRL. MGP-IRL extracted the tester heuristic in form of features and converted them to goals. We used MCTS and \replaced{Sarsa agents to play these test goals}{RL to play an agent according to its sequence of goals}. These goals directed the agent to different states in the game and \added{agents} generated test sequences. These sequences are executed on the game while the oracle checks the game to determine whether the game behaves as expected.

Our results show that the interaction state assisted capturing the human tester heuristic even if the game had bugs and supported MCTS and \replaced{Sarsa}{RL} to play the game as testers. The synthetic agent surpassed the baseline agent \deleted{---which only covered the game scenario graph--- }and most of the individual human-testers and human-like agents. Furthermore, human-like agents, when act together, can compete with the performance of the synthetic agent as well as that of human testers.

We also investigated the bug finding performance of MCTS and \replaced{Sarsa. We}{RL, and} found that the mean \replaced{performance of Sarsa is better than MCTS, but}{performances were similar and} the stochasticity of MCTS is useful in testing.\added{ Furthermore, the combined scores of all 5 MCTS runs showed that human-like MCTS agent competes with humans.} Lastly, due to our MGP-IRL algorithm, human-like agent with likelihood threshold of 0.0 behaved similar to the human testers.

To the best of our knowledge, this study is the first to propose\deleted{ the} human-like tester agents. We showed that these agents can successfully test unexplored levels. Besides, synthetic agent takes model based testing further by introducing the generally acknowledged agent concept in gaming to traditional test techniques.\added{ Moreover, once these agents are created, they can test a game any number of times, decreasing the human test effort.} Finally, proposing an interaction state enables us to catch the tester \replaced{strategies}{instincts} and play accordingly.

In the future, we would like to \replaced{use}{explore using} function approximators in RL\added{ agents}. Besides\added{,} we would like to implement an MGP-IRL that is more robust to random actions. Lastly, we would like to generalize this concept to 3D games and investigate how to model an interaction state in this environment.

\section*{Acknowledgment}
The authors would like to thank the testers participated in our experimentation.

\bibliographystyle{IEEEtran}
\bibliography{IEEEabrv,References}

\onecolumn
\appendices

\section{Realized Nodes of Game Scenario Graph}
\label{sec:appendix_1}

\clearpage
\section{Games}
\label{sec:appendix_2}

\begin{figure}[!ht]
  \centering
  \subcaptionbox{Game A (6x7) Level 1\label{fig:game_6x7_1}}
      {\includegraphics[width=0.48\columnwidth]{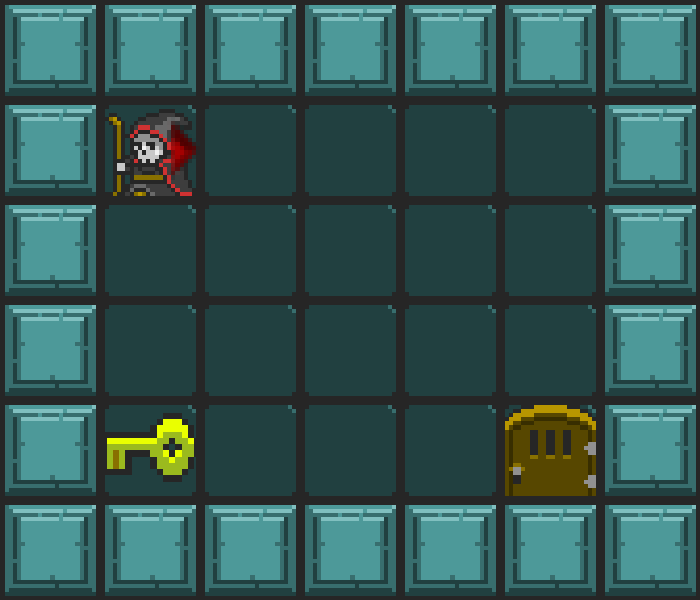}} \hfill
  \subcaptionbox{Game A (6x7) Level 2\label{fig:game_6x7_2}}
      {\includegraphics[width=0.48\columnwidth]{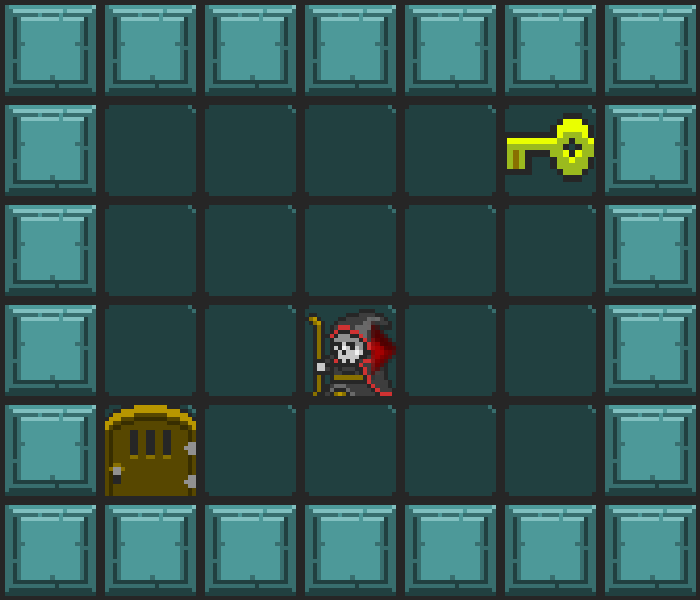}} \hfill
  \subcaptionbox{Game A (6x7) Level 3\label{fig:game_6x7_3}}
      {\includegraphics[width=0.48\columnwidth]{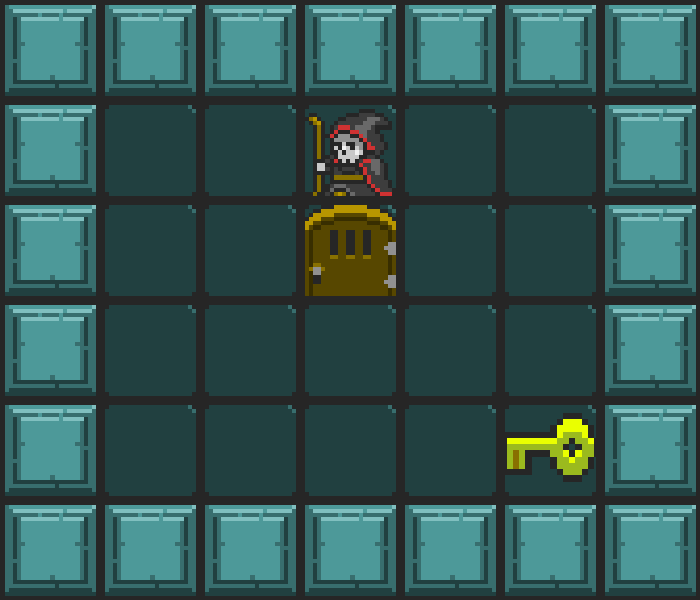}} \hfill
  \subcaptionbox{Game A (6x7) Level 4\label{fig:game_6x7_4}}
      {\includegraphics[width=0.48\columnwidth]{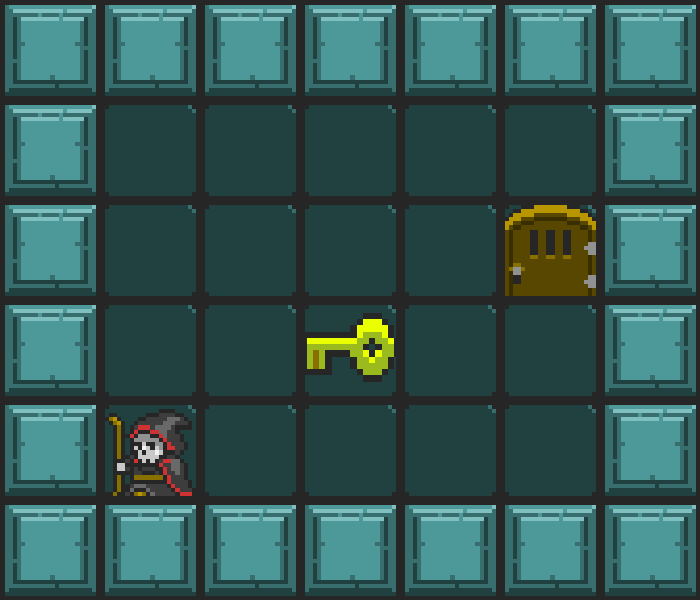}} \hfill
  \caption{Game A (6x7) Levels, Representing the Start of the Game}
  \label{fig:game_a_levels}
\end{figure}

\begin{figure}[]
  \centering
  \subcaptionbox{Game B (8x9) Level 1\label{fig:game_8x9_1}}
      {\includegraphics[width=0.48\columnwidth]{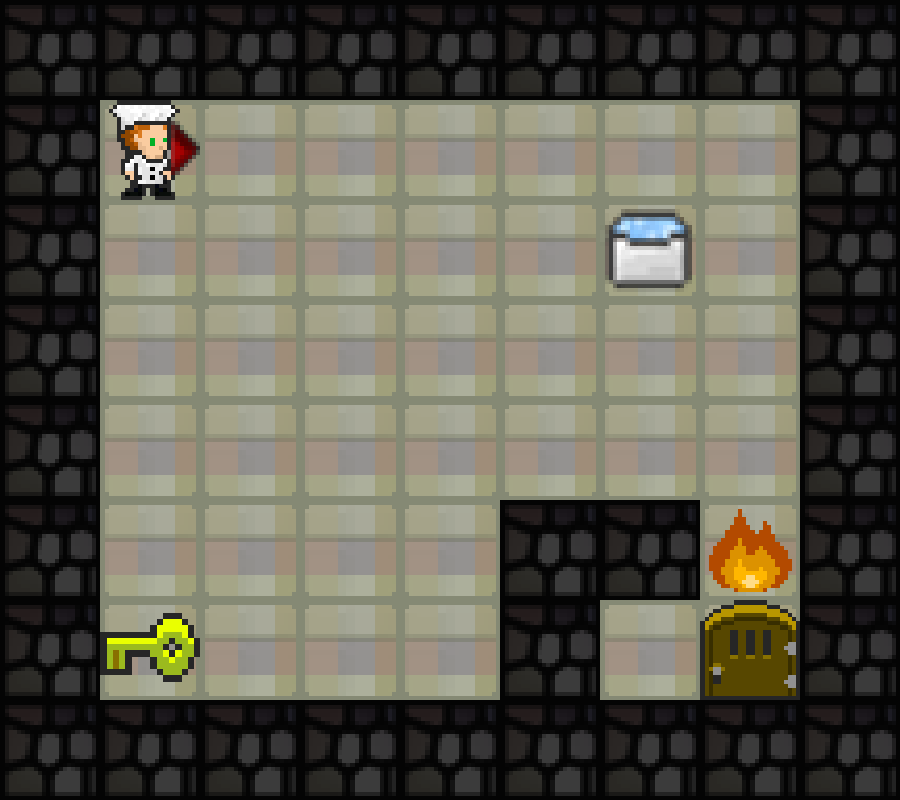}} \hfill
  \subcaptionbox{Game B (8x9) Level 2\label{fig:game_8x9_2}}
      {\includegraphics[width=0.48\columnwidth]{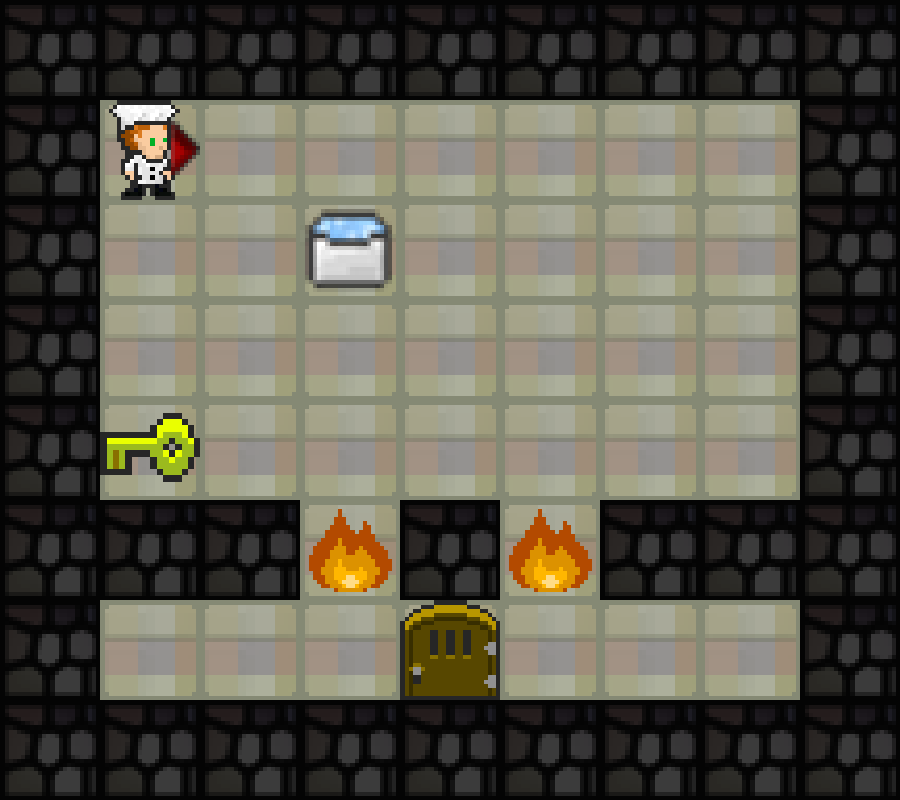}} \hfill
  \subcaptionbox{Game B (8x9) Level 3\label{fig:game_8x9_3}}
      {\includegraphics[width=0.48\columnwidth]{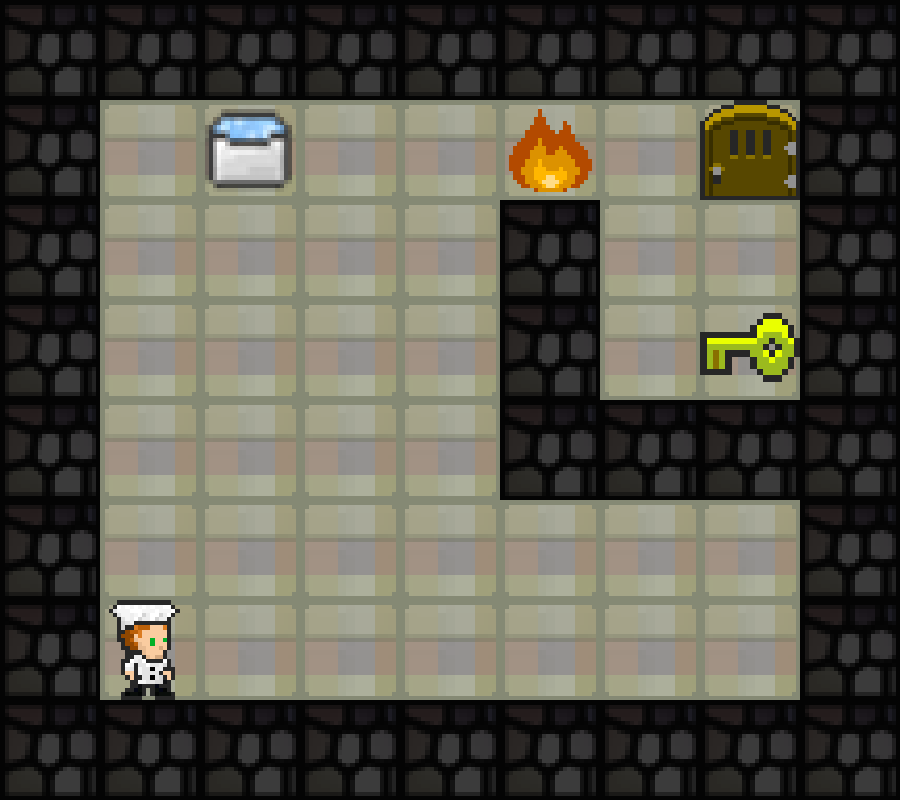}} \hfill
  \subcaptionbox{Game B (8x9) Level 4\label{fig:game_8x9_4}}
      {\includegraphics[width=0.48\columnwidth]{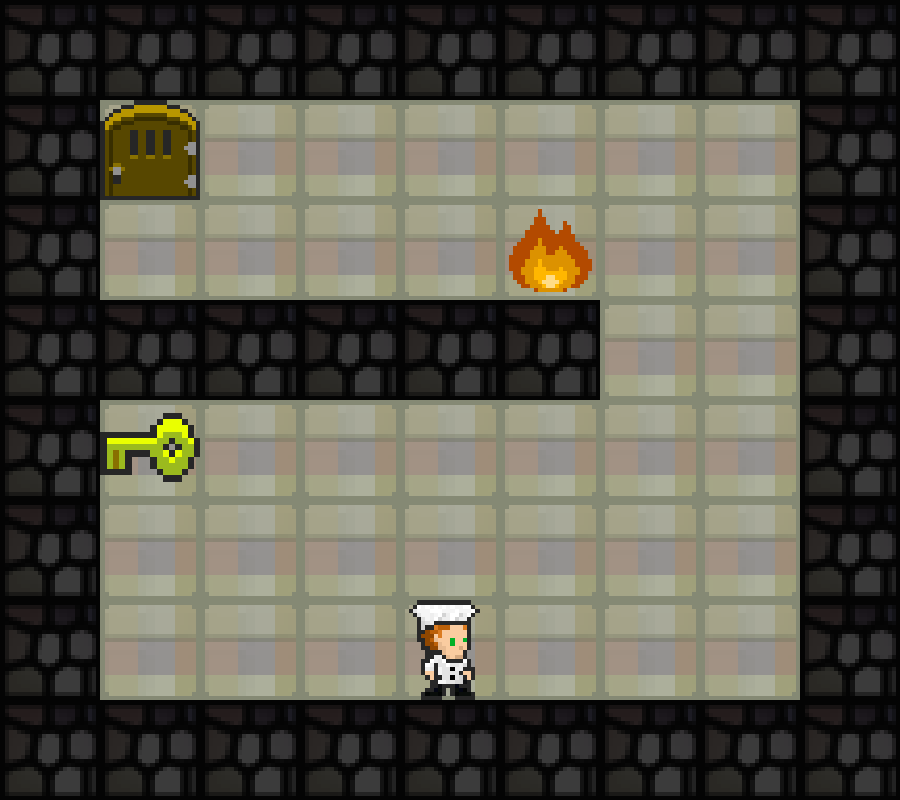}} \hfill
  \caption{Game B (8x9) Levels, Representing the Start of the Game}
  \label{fig:game_b_levels}
\end{figure}

\begin{figure}[]
  \centering
  \subcaptionbox{Game C (10x11) Level 1\label{fig:game_10x11_1}}
      {\includegraphics[width=0.48\columnwidth]{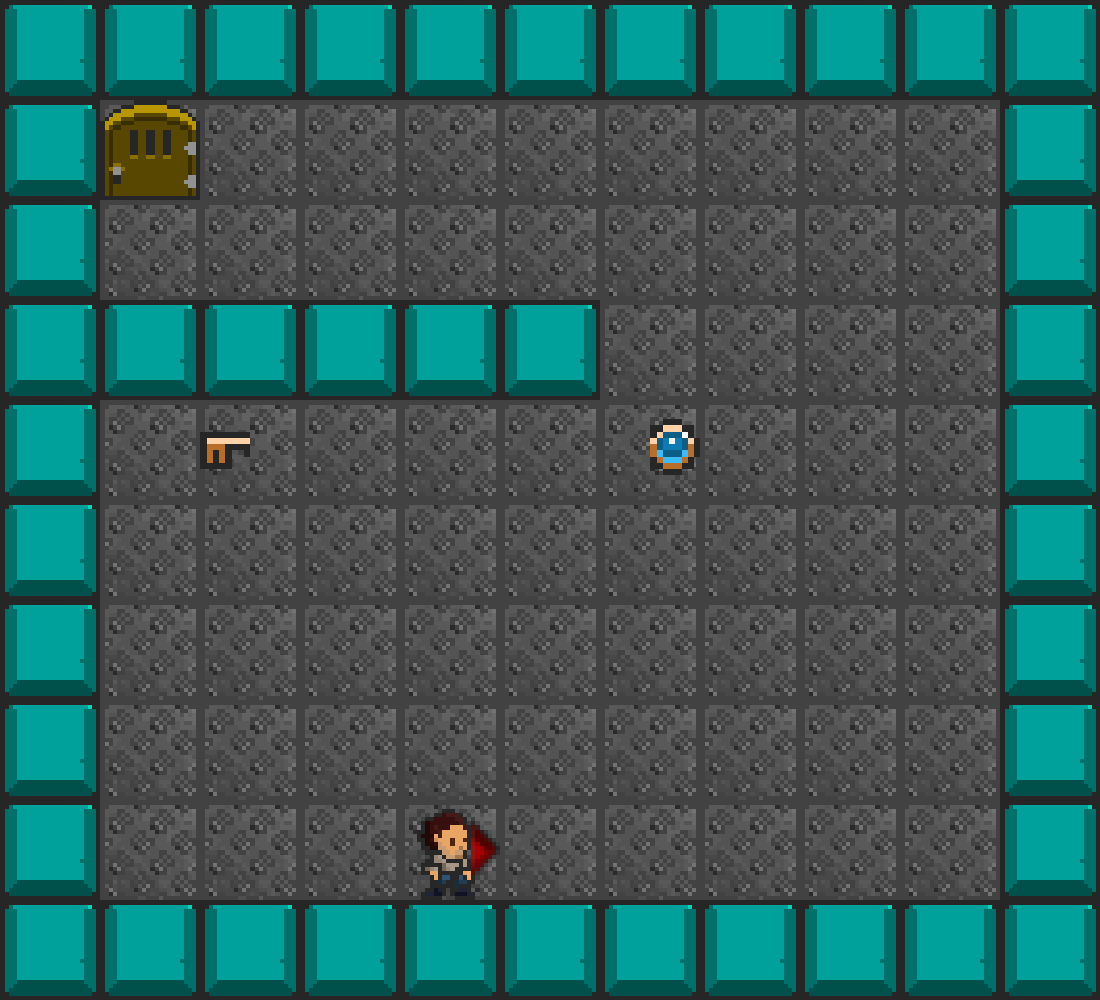}} \hfill
  \subcaptionbox{Game C (10x11) Level 2\label{fig:game_10x11_2}}
      {\includegraphics[width=0.48\columnwidth]{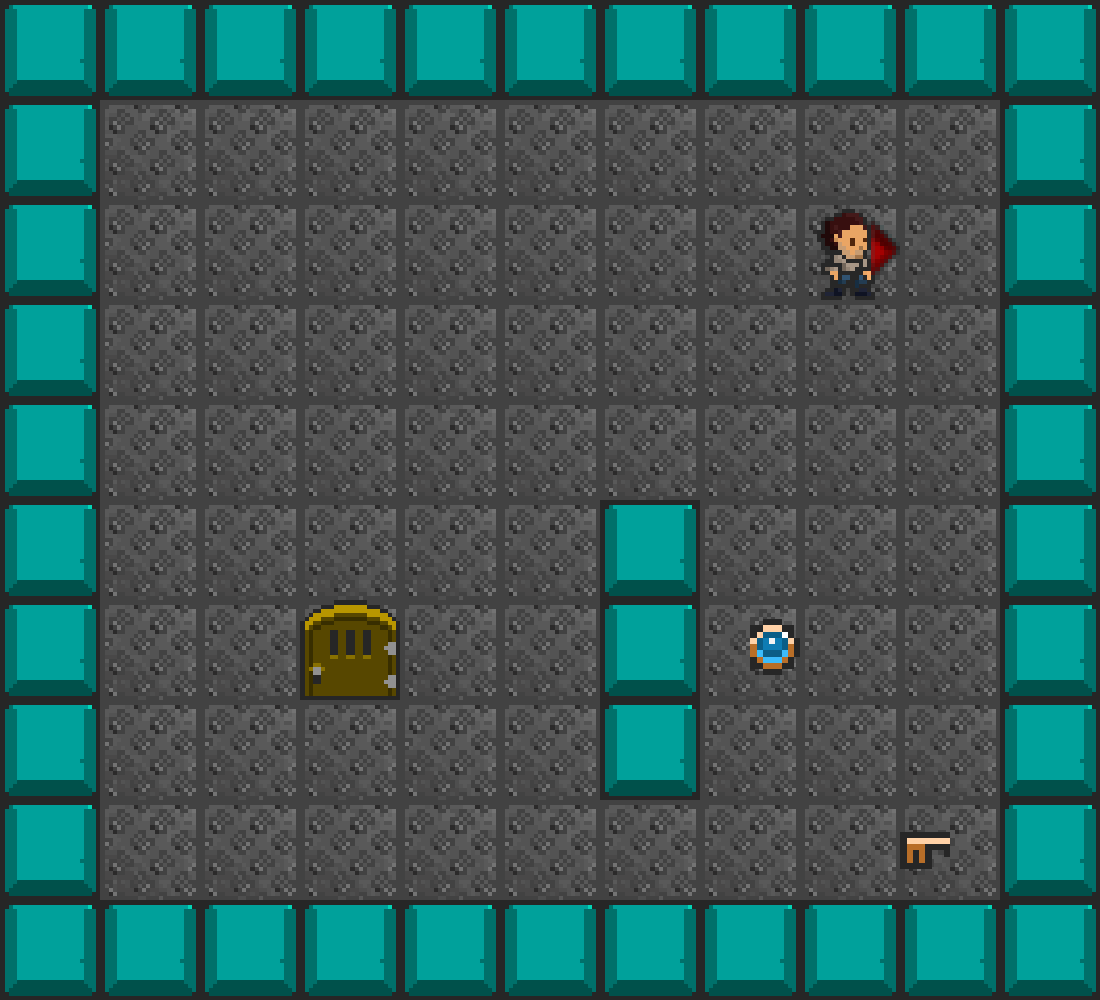}} \hfill
  \subcaptionbox{Game C (10x11) Level 3\label{fig:game_10x11_3}}
      {\includegraphics[width=0.48\columnwidth]{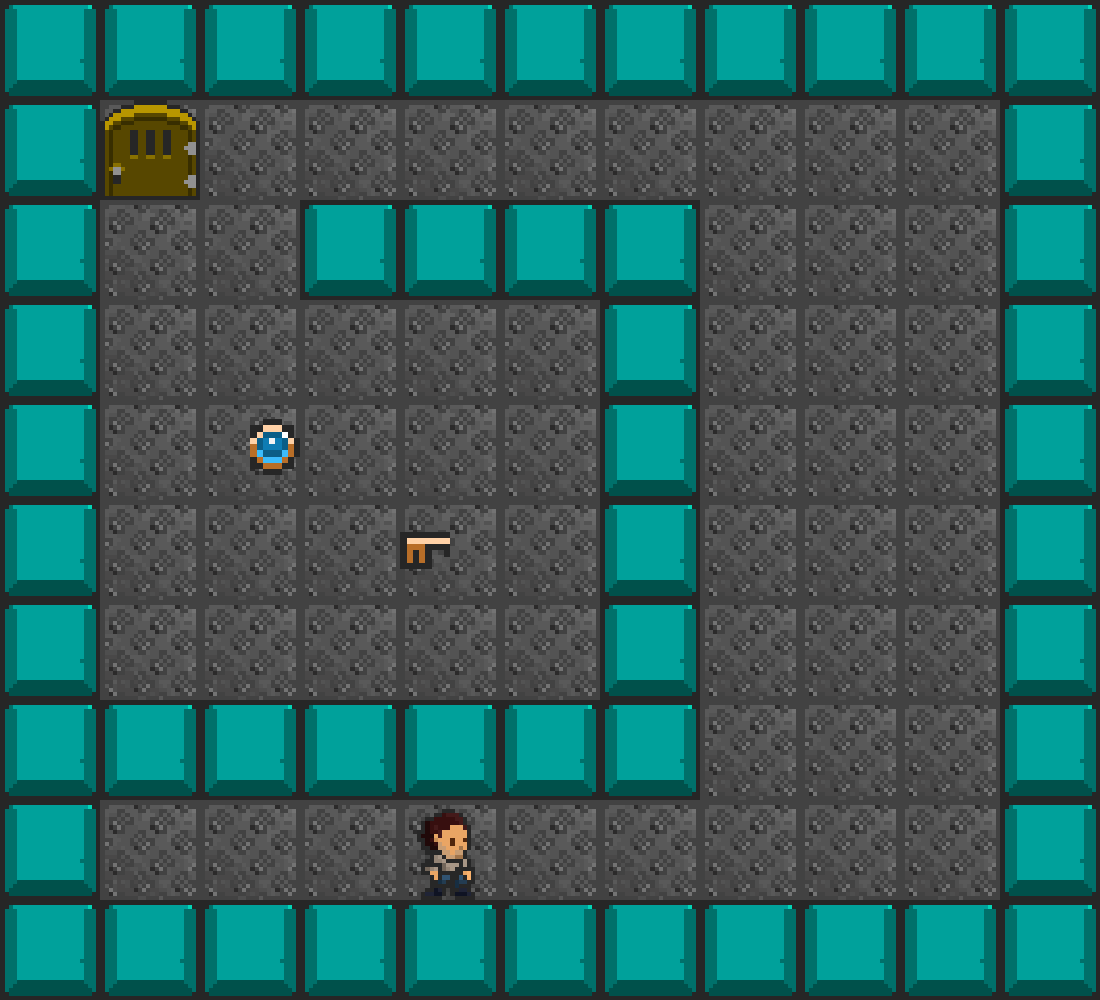}} \hfill
  \subcaptionbox{Game C (9x11) Level 4\label{fig:game_10x11_4}}
      {\includegraphics[width=0.48\columnwidth]{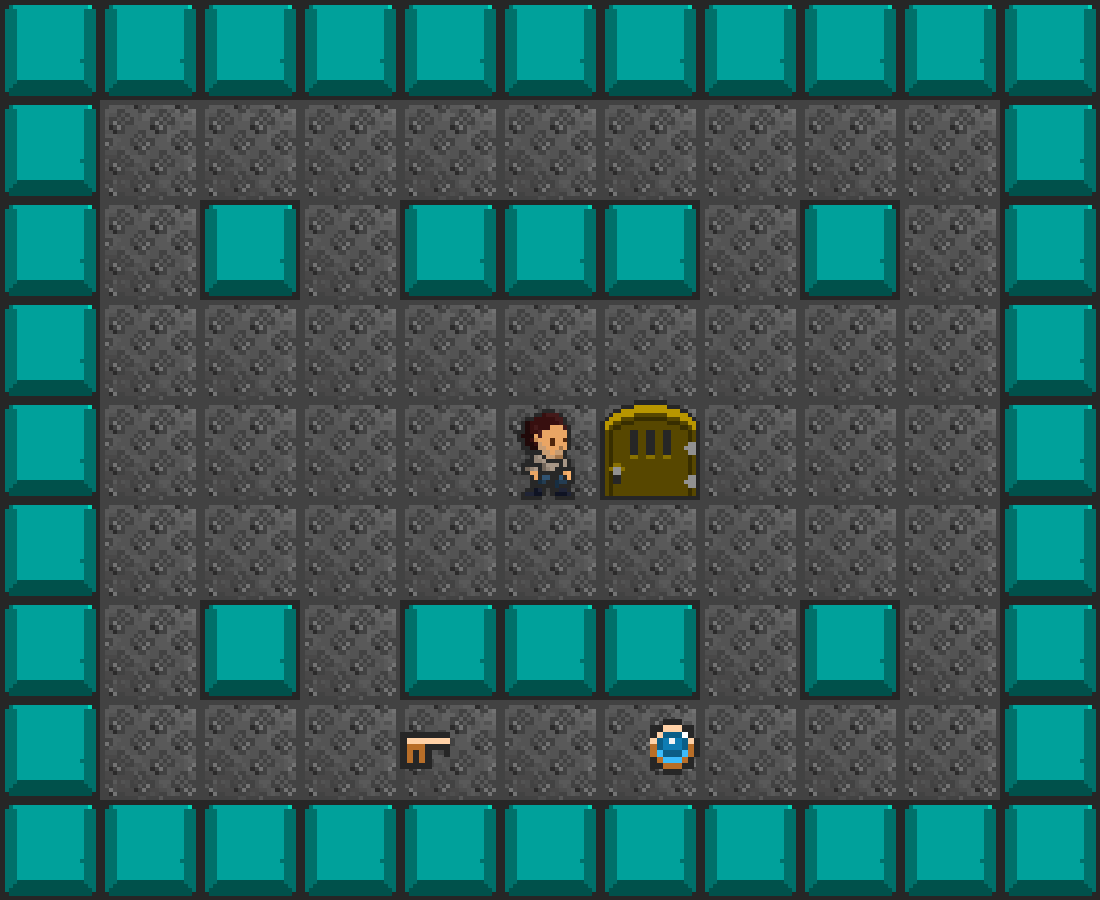}} \hfill
  \caption{Game A (10x11) Levels, Representing the Start of the Game}
  \label{fig:game_c_levels}
\end{figure}

\clearpage
\section{Third Appendix: VGDL}
\label{sec:appendix_3}

\begin{figure}[!ht]
  \begin{footnotesize}
  \begin{Verbatim}[commandchars=\\\{\}]

    BasicGame square_size=60
    SpriteSet
      floor > Immovable img=oryx/floor3 
      goal >
        goal2  > Door color=GREEN img=oryx/doorclosed1
        goal1  > Door color=GREEN img=oryx/doorclosed1
      key   > Immovable color=ORANGE img=oryx/key2
      sword >
        swordnokey > OrientedFlicker limit=9 singleton=True img=oryx/slash1
        swordkey > OrientedFlicker limit=9 singleton=True img=oryx/slash1
      avatar  > ShootAvatar
        nokey   > img=oryx/necromancer1 stype=swordnokey
        withkey > color=ORANGE img=oryx/necromancerkey1 stype=swordkey
      wall > Immovable autotiling=false img=oryx/wall3
  
    LevelMapping
      g > floor goal2
      + > floor key
      A > floor nokey
      w > floor wall
      . > floor
  
    InteractionSet
      avatar wall  > stepBack
      nokey goal2    > stepBack
      nokey goal1    > stepBack
      wall swordnokey > killSprite scoreChange=0
      goal1 swordkey > killSprite scoreChange=0
      goal2 swordkey > spawn stype=goal1 scoreChange=0
      goal2 swordkey > killBoth scoreChange=0
      goal2 withkey  > killSprite scoreChange=0
      nokey key     > transformTo stype=withkey scoreChange=0 killSecond=True
  
    TerminationSet
      SpriteCounter stype=goal   win=True
      SpriteCounter stype=avatar win=False

  \end{Verbatim}
  \end{footnotesize}
  \caption{VGDL of Game A (6x7) Level 1}
  \label{fig:vgdl_6x7_0}
\end{figure}

\begin{figure}[!ht]
  \begin{footnotesize}
  \begin{Verbatim}[commandchars=\\\{\}]

    BasicGame square_size=60
    SpriteSet
      floor > Immovable img=newset/floor6
      goal >
        goal2  > Door color=GREEN img=oryx/doorclosed1
        goal1  > Door color=GREEN img=oryx/doorclosed1
      key   > Immovable color=ORANGE img=oryx/key2
      sword >
        swordnokey > OrientedFlicker limit=9 singleton=True img=oryx/slash1
        swordkey > OrientedFlicker limit=9 singleton=True img=oryx/slash1
      debris > Immovable autotiling=false img=newset/whirlpool1
      avatar  > ShootAvatar
        nokey   > img=newset/chef stype=swordnokey
        withkey > color=ORANGE img=newset/chef_key stype=swordkey
      wall > Immovable autotiling=false img=newset/floor4
      water > Immovable autotiling=false img=oryx/water1
      fire > Passive autotiling=false img=oryx/fire1
  
    LevelMapping
      g > floor goal1
      + > floor key
      e > floor water
      f > floor fire
      A > floor nokey
      w > floor wall
      . > floor
  
  
    InteractionSet
      withkey water > killIfFromAboveNotMoving
      water avatar > bounceForward
      water wall goal key > undoAll
  
      avatar fire > killSprite
  
      fire water > transformTo stype=debris scoreChange=0 killSecond=True
      avatar wall  > stepBack
      nokey goal1    > stepBack
  
      goal1 withkey  > killSprite scoreChange=0
      nokey key     > transformTo stype=withkey scoreChange=0 killSecond=True
      water swordnokey > transformTo stype=fire  killSecond=True
  
  
    TerminationSet
      SpriteCounter stype=goal   win=True
      SpriteCounter stype=avatar win=False
  
  \end{Verbatim}
  \end{footnotesize}
  \caption{VGDL of Game B (8x9) Level 1}
  \label{fig:vgdl_8x9_0}
\end{figure}

\begin{figure}[!ht]
  \begin{footnotesize}
  \begin{Verbatim}[commandchars=\\\{\}]
    BasicGame square_size=60
    SpriteSet
      floor > Immovable img=newset/floor2
      goal > Door color=GREEN img=oryx/doorclosed1
      key   > Immovable color=ORANGE img=oryx/key3
      keyleft   > Immovable color=ORANGE img=oryx/key3_0
      keyright   > Immovable color=ORANGE img=oryx/key3_1
      sword > OrientedFlicker limit=9 singleton=True img=oryx/slash1
      avatar  > ShootAvatar
        nokey   > img=newset/man2 stype=sword
        withkey > color=ORANGE img=newset/man2_key stype=sword
      wall > 
        normalwall > Immovable autotiling=false img=newset/blockT
        fakewall > Immovable autotiling=false img=newset/blockT
  
    LevelMapping
      g > floor goal
      r > floor keyright
      l > floor keyleft
      A > floor nokey
      w > floor normalwall
      t > floor fakewall
      . > floor
  
    InteractionSet
      keyleft avatar > bounceForward
      keyright avatar > bounceForward
      keyleft goal normalwall > undoAll
      keyright goal normalwall fakewall > undoAll
  
      keyleft keyright > transformTo stype=key killSecond=true
      avatar wall > stepBack
  
      key sword > killSprite
      nokey goal > stepBack
      goal withkey  > killSprite scoreChange=0
      nokey key     > transformTo stype=withkey scoreChange=0 killSecond=true
  
    TerminationSet
      SpriteCounter stype=goal   win=True
      SpriteCounter stype=avatar win=False
  
  \end{Verbatim}
  \end{footnotesize}
  \caption{VGDL of Game C (10x11) Level 1.}
  \label{fig:vgdl_10x11_0}
\end{figure}

\end{document}